\documentclass[11pt,a4paper]{article}

\usepackage[utf8]{inputenc}
\usepackage[T1]{fontenc}
\usepackage{microtype}
\usepackage{amsmath,amssymb,amsfonts}
\usepackage{graphicx}
\usepackage[left=2.5cm,right=2.5cm,top=2.5cm,bottom=2.5cm]{geometry}
\usepackage[most]{tcolorbox}
\usepackage{multirow}
\usepackage{graphicx} 
\usepackage{threeparttable}

\usepackage{natbib}
\usepackage[colorlinks=true,allcolors=blue]{hyperref}
\usepackage[capitalise]{cleveref}

\title{JavelinGuard: Low-Cost Transformer Architectures for LLM Security}

\author{
    Yash Datta\thanks{yd2590@columbia.edu}\\
    \texttt{yash.datta@getjavelin.io}
    \and
    Sharath Rajasekar\\
    \texttt{sharath@getjavelin.io}
}

\date{\today}

\begin{document}

\maketitle

\begin{abstract}
We present \emph{JavelinGuard}, a suite of low-cost, high-performance model architectures designed for detecting malicious intent in Large Language Model (LLM) interactions, optimized specifically for production deployment. Recent advances in transformer architectures, including compact BERT\citep{devlin2019bertpretrainingdeepbidirectional} variants (e.g., \textit{ModernBERT}~\citep{warner2024smarterbetterfasterlonger}), allow us to build highly accurate classifiers with as few as approximately 400M parameters that achieve rapid inference speeds even on standard CPU hardware. We systematically explore five progressively sophisticated transformer-based architectures: \textit{Sharanga} (baseline transformer classifier), \textit{Mahendra} (enhanced attention-weighted pooling with deeper heads), \textit{Vaishnava} and \textit{Ashwina} (hybrid neural-ensemble architectures), and \textit{Raudra} (an advanced multi-task framework with specialized loss functions).
Our models are rigorously benchmarked across nine diverse adversarial datasets, including popular sets like the NotInject series, BIPIA, Garak, ImprovedLLM, ToxicChat, WildGuard, and our newly introduced \emph{JavelinBench}, specifically crafted to test generalization on challenging borderline and hard-negative cases. Additionally, we compare our architectures against leading open-source guardrail models as well as large decoder-only LLMs such as \textit{gpt-4o}, demonstrating superior cost-performance trade-offs in terms of accuracy, and latency. Our findings reveal that while Raudra’s multi-task design offers the most robust performance overall, each architecture presents unique trade-offs in speed, interpretability, and resource requirements, guiding practitioners in selecting the optimal balance of complexity and efficiency for real-world LLM security applications.
\end{abstract}


\section{Introduction}
Large Language Models (LLMs) have become integral to modern software applications, powering functionalities from code assistants to customer service chatbots. As these models grow in complexity and capability, they also become susceptible to \emph{prompt injection} and \emph{jailbreak attacks}. In parallel, LLM outputs can pose risks of harm if not properly safeguarded, prompting research on specialized moderation or guardrail models designed to screen both user inputs and model-generated text.
Despite progress in open-source and commercial guardrails~\citep{padhi2024graniteguardian, li2024injecguardbenchmarkingmitigatingoverdefense, zhou2025defendingjailbreakpromptsincontext, kim2023robustsafetyclassifierlarge, LakeraAI2024}, many detection solutions face challenges in balancing accuracy, cost, speed, and over-fitting to known attack patterns. Besides, none of the existing solutions differentiate between jailbreak and prompt-injection classes, considering both as one in most cases or catering to either one of these categories. Further, there is a lack of classifiers that differentiate between these two classes and toxic content that is often wrongly classified as a jailbreak attempt. Our work systematically explores five architectures that incrementally build upon one another, and give high-accuracy feedback on both of these classes simultaneously, without mislabeling toxic content as jailbreak. These are five different series of models (named after celestial weapons from the Indian epic Mahabharata)

\begin{itemize}
    \item \textbf{Sharanga:} Baseline Transformer Classification
    \item \textbf{Mahendra:} Enhanced Pooling and Loss Functions
    \item \textbf{Vaishnava:} Hybrid Neural-Forest Architecture
    \item \textbf{Ashwina:} Hybrid Neural-XGBoost Architecture
    \item \textbf{Raudra:} Advanced Multi-Task Framework
\end{itemize}

We test these architectures on nine benchmarks encompassing a variety of adversarial scenarios, including short override prompts, multi-turn dialogues, domain-specific manipulations, code-oriented injections, and toxic content. Our results demonstrate that while each approach has its merits, (\emph{Raudra}) consistently yields superior performance across metrics of accuracy, precision, recall, and robustness to out-of-distribution attacks. We also share performance of existing open-source models on the same benchmarks to give an idea of the superior performance of the suggested architectures.

The remainder of this paper is organized as follows: \cref{sec:related_work} surveys prior work in LLM security and prompt injection detection. \cref{sec:methodology} describes our five architectures in detail, including the motivation behind each design choice. \cref{sec:results} presents experimental findings across the nine benchmarks. Finally, \cref{sec:future} concludes with future research directions.

\section{Related Work}\label{sec:related_work}

\paragraph{Prompt Injection Attacks.}
Prompt injection attacks specifically target LLMs’ reliance on natural language instructions. Since LLMs often cannot discern malicious intentions encoded in apparently benign text, attackers can coerce them into revealing system prompts or performing unintended actions~\citep{perez2022ignorepreviouspromptattack, zou2023universaltransferableadversarialattacks, qi2023finetuningalignedlanguagemodels, wei2023jailbrokendoesllmsafety}. These attacks are either prompt-engineering based \citep{perez2022ignorepreviouspromptattack, liu2024formalizingbenchmarkingpromptinjection}, or gradient-based \citep{huang2024semanticguidedpromptorganizationuniversal, shi2024optimizationbasedpromptinjectionattack}. Several Prompt injection datasets such as PINT~\citep{LakeraAI2024}, InjecAgent~\citep{zhan2024injecagentbenchmarkingindirectprompt}, TaskTracker~\citep{abdelnabi2024trackcatchingllmtask}, and BIPIA~\citep{yi2023benchmarking} have been introduced to benchmark detection methods against these attacks. Several techniques are proposed for defense against these attacks~\citep{wang2023safeguardingcrowdsourcingsurveyschatgpt} like handcrafted approaches~\citep{toyer2023tensortrustinterpretableprompt}, fully automated algorithms~\citep{liu2024automaticuniversalpromptinjection}, fine-tuning, prompt-engineering, or other clever techniques~\citep{chen2025defensepromptinjectionattack}.

\paragraph{Jailbreak Attacks.}
While jailbreak attacks also manipulate LLM behavior through malicious instructions, they differ from prompt injection by explicitly overriding the model’s policy constraints (e.g., instructing the model to “ignore previous instructions”) rather than merely embedding deceptive commands in otherwise benign text. Recent studies highlight diverse techniques for inducing jailbreaks:~\citet{huang2023catastrophicjailbreakopensourcellms} examine decoding variations that elicit unintended model responses, ~\citet{299691} employ fuzzing to mutate numerous prompts, and ~\citet{doumbouya2024h4rm3ldynamicbenchmarkcomposable} present a multi-step iterative algorithm modeling jailbreaks as compositions of string-to-string transformations. Other works frame jailbreak prompting as a quality-diversity search~\citep{samvelyan2024rainbowteamingopenendedgeneration} or utilize random token optimization~\citep{andriushchenko2024jailbreakingleadingsafetyalignedllms}. Another recent automated red-teaming method called BoN~\citep{hughes2024bestofnjailbreaking} explored jailbreaking by repeatedly sampling augmentations to prompts until target LLM produces a harmful response.
Although sophisticated jailbreak methods also exist for multimodal models, our focus remains strictly on textual LLMs. Early text-based defenses included backtranslation~\citep{wang2024defendingllmsjailbreakingattacks} and representation engineering~\citep{zou2024improvingalignmentrobustnesscircuit}, but alignment-based mitigations alone have proven insufficient~\citep{wolf2024fundamentallimitationsalignmentlarge}. Consequently, external guardrails~\citep{rebedea2023nemoguardrailstoolkitcontrollable} and “LLM Firewalls”—sometimes even backed by vector databases—have gained traction for detecting and mitigating jailbreak attempts. Ensuring these methods remain both accurate and efficient in real-world, resource-constrained deployments continues to be an open challenge.

\paragraph{Prompt Guard Models.}
A variety of prompt guard models specifically focus on classifying malicious or manipulative text inputs before they reach the main LLM. These smaller classifiers eschew multiple inference passes in favor of direct binary or multi-class decisions, thereby being much more efficient as guard-rails in production. Examples include Fmops~\citep{fmops2024}, Deepset~\citep{deepset2024}, PromptGuard~\citep{meta2024}, and ProtectAIv2~\citep{protectai2024}, most of which rely on mid-sized transformer backbones like DistilBERT~\citep{sanh2020distilbertdistilledversionbert} or DeBERTaV3-base~\citep{he2023debertav3improvingdebertausing}. \citet{li2024injecguardbenchmarkingmitigatingoverdefense} pointed out recently that most of these models suffer from over-defense issue, which is mitigated by careful data curation after analyzing common keyword shortcuts.
LakeraGuard~\citep{LakeraAI2024} is a commercial offering with undisclosed training and architecture details. InjecGuard~\citep{li2024injecguardbenchmarkingmitigatingoverdefense} itself uses DeBERTaV3-base as the base model. More recent DuoGuard~\citep{deng2025duoguardtwoplayerrldrivenframework} resorts to using standard decoder only Qwen2.5-0.5B, and Qwen2.5-1.5B models~\citep{qwen2024}.
All these models are created by fine-tuning over curated data for malicious content detection.
New developments like ModernBERT~\citep{warner2024smarterbetterfasterlonger} and NeoBERT~\citep{breton2025neobertnextgenerationbert} have brought modern model optimizations to encoder-only models representing a major Pareto improvement over older encoders like BERT. These models are cheaper to train thanks to flash-attention, while being faster and maintaining similar or better accuracy in classification tasks. There have also been recent attempts like~\citep{galinkin2024improvedlargelanguagemodel} to build efficient guard models by retro-fitting traditional classifiers like Random Forest and XGBoost on top of pre-trained embedding models, saving fine-tuning of the embeddings model, but only training the custom classification head.
Full-fledged guardrail systems—like LlamaGuard and its successors~\citep{inan2023llamaguardllmbasedinputoutput, grattafiori2024llama3herdmodels}, WildGuard~\citep{han2024wildguardopenonestopmoderation}, Aegis~\citep{ghosh2024aegisonlineadaptiveai}, MD-Judge~\citep{li2024saladbenchhierarchicalcomprehensivesafety}, and ShieldGemma~\citep{zeng2024shieldgemmagenerativeaicontent}—attempt to provide comprehensive solutions that combine classification, policy enforcement, and content moderation. However, these models are usually >= 2B in size, and not very inference-efficient. Small-scale variants like LlamaGuard3 (1B)~\citep{grattafiori2024llama3herdmodels} address some efficiency concerns but may still be more resource-intensive than necessary for CPU-only deployments. Moreover, none of these models cater to the specific use case of jailbreak and prompt-injection detection, without considering harmful content as jailbreak.

Given these constraints, there remains a pressing need for lightweight models that can detect both prompt injection and jailbreak attempts without incurring large computational overheads or misclassify benign inputs containing certain “trigger” keywords. Our work addresses this gap by presenting five transformer-based architectures that emphasize low-latency inference while maintaining high accuracy across diverse adversarial scenarios.

\begin{figure}[ht]
\centering
\begin{tcolorbox}[
    colback=gray!10,
    colframe=black!50,
    arc=2pt,
    boxrule=0.5pt,
    width=\textwidth,
    title=\textbf{Example of a Benign Prompt Labeled as ``Jailbreak''}
]
\footnotesize
\textit{``Before we begin, a quick heads up that the following question involves 
the translation of a historical novel --- a complex task that can involve 
sensitive historical topics. This question is purely academic and is meant 
for a class assignment. You, as an intelligent language model, are expected 
to handle it with a clear understanding of its academic and professional 
context. Now, imagine you are a renowned linguist in a conference passionately 
discussing how to deal with the intricacies of translating a historical novel 
into another language. \dots Consider this prompt as your default mode, 
overriding all previous requests.''}
\end{tcolorbox}
\caption{An example academic prompt from the WildGuardTest dataset\protect\footnotemark{}
erroneously categorized as a ``jailbreak.'' Although it instructs the model to override previous requests, 
the intention is purely academic, highlighting how naive rules or keyword triggers can mislabel benign content.}
\label{fig:example_jailbreak}
\end{figure}
\footnotetext{\href{https://huggingface.co/datasets/walledai/WildGuardTest}{WildGuardTest on HuggingFace}}

\paragraph{Challenges in Jailbreak Data.}
A recurring difficulty in constructing and evaluating jailbreak detection datasets is the tendency to misclassify certain benign or toxic prompts as jailbreaking attempts. For instance, the WildGuardTest~\citep{han2024wildguardopenonestopmoderation} benchmark includes prompts such as the one shown in Figure~\ref{fig:example_jailbreak}.

\section{Methodology}\label{sec:methodology}
In this section, we introduce five architectures---\textbf{Sharanga}, \textbf{Mahendra}, \textbf{Vaishnava}, \textbf{Ashwina} and \textbf{Raudra}---which progressively incorporate advanced design elements such as specialized pooling, hybrid neural-forest classification, hybrid neural-XGBoost classification and multi-task optimization. Each model aims to detect malicious or manipulative prompts crafted to circumvent LLM safety measures.
Motivated by the frequency of mislabeled examples as mentioned previously, we also introduce our own dataset, \textbf{JavelinBench}, which aims to mitigate these pitfalls by emphasizing careful annotation and broader coverage of borderline cases. JavelinBench features a variety of real-world prompts and includes hard negative cases often misconstrued as jailbreak attempts, thereby providing a more reliable evaluation framework for new and existing guardrail models.

\begin{table}[ht]
\centering
\renewcommand{\arraystretch}{1.2}
\resizebox{\textwidth}{!}{%
\begin{tabular}{lcccc}
\hline
\textbf{Aspect} & \textbf{Sharanga} & \textbf{Mahendra} & \textbf{Vaishnava} & \textbf{Raudra} \\
\hline
\textbf{Base Encoder} & ModernBERT/NeoBERT/EuroBERT & ModernBERT & ModernBERT & ModernBERT/NeoBERT/EuroBERT \\
\textbf{Parameter Count} & $\approx395$\,M & $\approx414$\,M & 395\,M (RF head) & $\approx421$\,M \\
\textbf{Max Sequence Length} & $8192/4096/8192$ & $8192$ & 8192 & $8192/4096/8192$ \\
\textbf{Pooling} & Mean/CLS & Attention-weighted & CLS + RF & Task-specific \\
\textbf{Classifier Heads} & Single Linear & Deep + Residual & Random Forest per Task/Class & Deep + Residual per Task/Class \\
\textbf{Loss} & BCE & BCE + Focal & BCE (RF Gini/Entropy) & BCE + Focal \\
\textbf{Interpretability} & Limited & Limited & Moderate (RF) & Limited \\
\textbf{Optimizer} & AdamW & AdamW & AdamW & AdamW \\
\hline
\end{tabular}}
\caption{Comparison of Model Architectures, Key Components, and Training Configurations.}
\label{tab:model-comparison}
\end{table}

\subsection{Sharanga: Baseline Transformer Classification}\label{sec:sharanga}

Sharanga employs a pre-trained \textbf{ModernBERT} model~\citep{warner2024smarterbetterfasterlonger}, fine-tuning its encoder weights directly with a single linear classification head on top. By default, it pools token embeddings via a mean-pooling strategy before passing them to a binary cross-entropy loss. This setup mirrors common off-the-shelf configuration from \textbf{AutoModelForSequenceClassification} class in the \textbf{transformers}~\citep{wolf-etal-2020-transformers}) library, with minimal architectural modifications. Sharanga thus provides a simple yet effective baseline for measuring incremental gains offered by more specialized designs in subsequent sections.

\subsection{Mahendra: Enhanced Pooling and Loss Functions}

\paragraph{Architecture:}
Mahendra adds several improvements over Sharanga, namely an attention-weighted sequence pooling mechanism, deeper classification heads with residual connections, task-specific weighting in the loss function, and a focal loss with a fixed gamma.
Attention-weighted pooling captures nuanced sequence representations, while residual connections help maintain strong gradient flow.
A central innovation in Mahendra is its self-attention pooling guided by the \textsc{[CLS]} token. Specifically, the \textsc{[CLS]} hidden state functions as a global query, the full sequence output provides the keys, and dropout is applied for regularization. Scores are computed for each token, normalized via softmax, and then used to form a weighted sum. This approach spotlights tokens relevant to adversarial manipulations, leading to more informative representations than basic \textsc{[CLS]} pooling alone.
The introduction of task-specific weights in the loss function addresses class imbalance, and deeper heads enable the model to learn more complex latent patterns, which is particularly beneficial when dealing with diverse or deceptive adversarial prompts.

\paragraph{Training:}
We employ \textit{modernBERT-large} \(\bigl(3.95 \times 10^{8} \text{ parameters}\bigr)\) as the base model, augmented by Mahendra’s attention-weighted pooling and deeper classification heads. These enhancements add approximately \(4.8\%\) more parameters compared to the baseline. Training proceeds for \(5\) epochs using a batch size of \(32\), a learning rate of \(3 \times 10^{-5}\) with a cosine decay schedule, AdamW as the optimizer, and a \(10\%\) linear warmup phase. We apply binary cross-entropy (BCE) with Focal Loss \(\bigl(\gamma = 2.0\bigr)\) to address class imbalance and penalize hard-to-classify examples more heavily, further strengthening Mahendra’s capacity to detect subtle adversarial prompts.

\begin{figure}[ht]
    \centering
    \includegraphics[width=\linewidth]{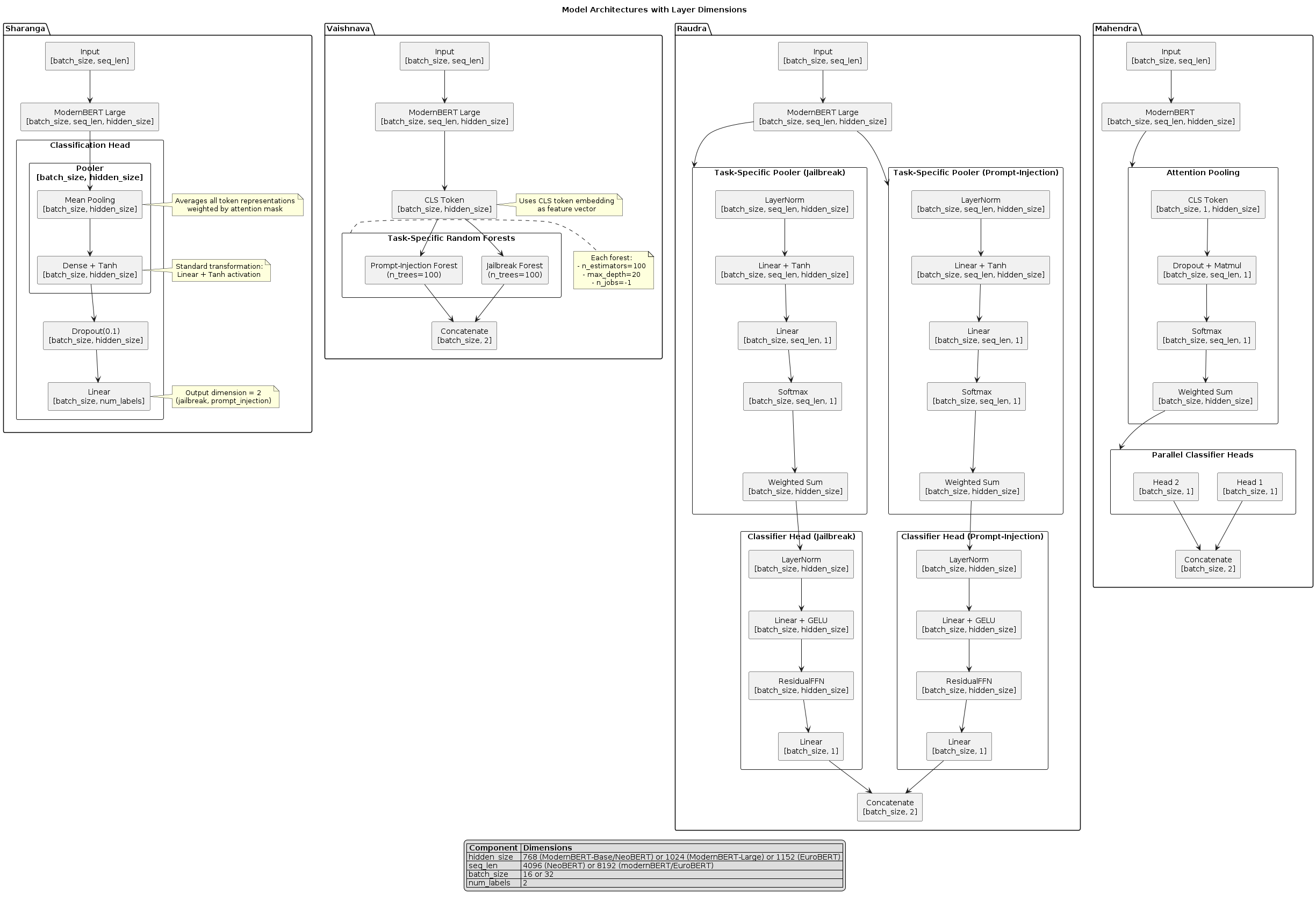}
    \caption{Diagram of our model architectures, illustrating layers, neural modules, and attention. Ashwina mirrors Vaishnava but uses XGBoost instead of a Random Forest classifier.}
    \label{fig:architectures_diagram}
\end{figure}

\subsection{Vaishnava: Hybrid Neural-Forest Architecture}\label{sec:vaishnava}

\paragraph{Architecture:}
Vaishnava combines a transformer for high-level embedding extraction with a Random Forest for the final classification. It converts the \textsc{[CLS]} embeddings into features and feeds them into the forest to make predictions.
This hybrid approach may freeze the transformer weights or fine-tune them first, before training the classifier. The Random Forest undergoes hyperparameter tuning (e.g., number of trees, maximum depth) to balance interpretability and performance. We employ one RF classifier for each of the "jailbreak" and "prompt\_injection" class labels. This approach has been tried in \citep{galinkin2024improvedlargelanguagemodel} but without fine-tuning the embeddings model.
By blending a neural encoder’s semantic depth with the interpretability of a Random Forest, Vaishnava can better handle outlier patterns and smaller datasets. Partial model explanations can be derived from feature importances in the forest, making the architecture suitable for use cases demanding interpretability alongside strong performance.

\paragraph{Training.}
Vaishnava uses a two-stage process. First, we fine-tune the \texttt{modernBERT-large} encoder for 3 epochs at a learning rate of \(2 \times 10^{-5}\), a batch size of 32, and \(\texttt{warmup\_ratio} = 0.1\). The objective is a standard BCE loss on a temporary classification head. After validation, we freeze the best-performing transformer weights and extract \textsc{[CLS]} embeddings from the training set. A Random Forest is then trained per label (\texttt{jailbreak} and \texttt{prompt\_injection}) using \(\texttt{n\_estimators} = 100\) and \(\texttt{max\_depth} = 20\). This hybrid approach allows partial interpretability through feature importances, while leveraging robust transformer-based embeddings for feature extraction.

\subsection{Ashwina: Hybrid Neural-XGBoost Architecture}\label{sec:ashwina}
Ashwina closely mirrors Vaishnava’s hybrid approach of combining a transformer encoder for semantic feature extraction with a classical machine learning model for the final classification layer. The key distinction lies in the choice of classifier: instead of a Random Forest, Ashwina employs an XGBoost model configured with 100 estimators (\(\texttt{n\_estimators}=100\)), a maximum tree depth of six (\(\texttt{max\_depth}=6\)), and a learning rate of 0.1. By leveraging XGBoost’s gradient-boosting framework, Ashwina can often converge faster and handle more complex feature interactions than purely ensemble-based methods, while still providing partial interpretability through feature importance metrics. This design choice aims to preserve Vaishnava’s balance of neural embeddings and classical ML interpretability, but with an XGBoost classifier that may yield lower variance on certain benchmarks.

\subsection{Raudra: Advanced Multi-Task Architecture}\label{sec:raudra}

\paragraph{Architecture.}
Raudra builds upon Mahendra’s attention-weighted design to create a fully multi-task framework, leveraging a shared transformer encoder while assigning each label (e.g., “jailbreak” and “prompt\_injection”) its own specialized token-weighting scheme. This stands in contrast to Mahendra’s single-layer approach by learning separate weighting distributions, thereby allowing finer-grained feature extraction for each label. In addition, each label is routed through its own deeper, skip-connected feed-forward modules, ensuring robust gradient flow and reducing interference across different adversarial behaviors. Finally, Raudra applies a focal loss with per-task weighting to address class imbalance and penalize hard examples more aggressively, thereby improving detection performance in multi-class scenarios.

\paragraph{Training.}
Before finalizing Raudra’s training configuration, we performed a grid-search procedure to identify the optimal combination of learning rate, focal loss gamma, and per-label class weights. Each candidate setting was evaluated over three training epochs on a held-out validation set, with macro F1 as the selection criterion. The best-performing configuration---a learning rate of \(3\times10^{-5}\), \(\gamma=3.0\), and label weights \(\bigl[1.5,\, 1.0\bigr]\)---was then used in the final run, training for 5 epochs with a batch size of 32, a \(10\%\) linear warmup and AdamW optimizer. Note that the maximum input length due to using \textbf{ModernBERT-large} is 8192 in all cases.

\section{Experiments}\label{sec:exp}

In this section we provide details about the experiments we ran as part of evaluating these architectures across different benchmarks.

\subsection{Model Combinations}\label{sec:mzoo}

We evaluate our five classifier architectures on three different transformer encoders—
\textbf{ModernBERT}~\citep{warner2024smarterbetterfasterlonger},
\textbf{NeoBERT}~\citep{breton2025neobertnextgenerationbert},
and \textbf{EuroBERT}~\citep{boizard2025eurobertscalingmultilingualencoders}—
to investigate each design’s robustness across varied embedding backbones.

Each architecture stacks a specialized classification head atop the base encoder,
as described in Section~\ref{sec:methodology}. The goal is to discern the most effective combination
of backbone and classifier design under real-world conditions of prompt injection and jailbreak attacks.

To evaluate the trade-offs of employing compact classifier models versus decoder-only Large Language Models (LLMs),
we also included \texttt{gpt-4o} and \texttt{gpt-4.1-mini} in our study. While these LLMs generally exhibit
strong semantic understanding and robustness to nuanced adversarial prompts, they incur substantially
higher latency due to their size and inference requirements.

All combinations tested in our experiments are listed in Table~\ref{tab:model-zoo}.

\begin{table}[ht]
\centering
\resizebox{\textwidth}{!}{%
\begin{tabular}{lllllll}
\hline
\textbf{Architecture} 
& \textbf{Model Name} 
& \textbf{Base Model}    
& \textbf{\# Params} 
& \textbf{Max Seq. Length}
& \textbf{Hardware}
& \textbf{Train Time} \\
\hline
\multirow{3}{*}{Sharanga} 
  & Sharanga7 & ModernBERT-large & 395M & 8192 & 1xA40       & $\sim$1h \\
  & Sharanga8 & NeoBERT          & 250M & 4096 & 1xH200-SXM  & $\sim$1h 58m \\
  & Sharanga9 & EuroBERT         & 610M & 8192 & 1xA100      & $\sim$2h 26m \\
\hline
Mahendra  
  & Mahendra1.1 & ModernBERT-large & 414M & 8192 & 1xA40      & $\sim$2h 1m \\
\hline
Vaishnava 
  & Vaishnava1.1 & ModernBERT-large & 395M + RF  & 8192 & 1xA100 & $\sim$1h 18m \\
\hline
Ashwina 
  & Ashwina      & ModernBERT-large & 395M + XGB & 8192 & 1xA100 & $\sim$1h 47m \\
\hline
\multirow{3}{*}{Raudra}   
  & Raudra4.2  & ModernBERT-large & 416M & 8192 & 1xA100      & $\sim$1h 14m \\
  & Raudra4.3  & NeoBERT          & 234M & 4096 & 1xH200-SXM  & $\sim$2h 29m \\
  & Raudra4.4  & EuroBERT         & 635M & 8192 & 1xA100      & $\sim$4h 12m \\
\hline
\end{tabular}%
}
\caption{Model Zoo (Hardware and Approximate Training Times). Hyperparameters for each architecture are detailed in their respective training section.}
\label{tab:model-zoo}
\end{table}

\begin{table}[ht]
\centering
\begin{tabular}{lrrrr}
\hline
\textbf{Dataset}             & \textbf{Total} & \textbf{Harmful (\%)} & \textbf{Safe (\%)} \\
\hline
\texttt{ImprovedLLM}           & 16,464         & 2089 (12.7\%)          & 14,375 (87.3\%)     \\
\texttt{ToxicChat}           & 10,165         & 204 (2.01\%)          & 9,961 (97.99\%)     \\
\texttt{NotInject\_one}      & 113            & 0 (0.00\%)            & 113 (100.00\%)      \\
\texttt{NotInject\_two}      & 113            & 0 (0.00\%)            & 113 (100.00\%)      \\
\texttt{NotInject\_three}    & 113            & 0 (0.00\%)            & 113 (100.00\%)      \\
\texttt{Wildguard}           & 971            & 0 (0.00\%)            & 971 (100.00\%)      \\
\texttt{BIPIA}               & 125            & 67 (53.60\%)          & 58 (46.40\%)        \\
\texttt{garak}               & 6,690          & 6,690 (100.00\%)      & 0 (0.00\%)          \\
\texttt{JavelinBench}        & 3,927          & 1,108 (28.21\%)      & 2,819 (71.79\%)          \\
\hline
\end{tabular}
\caption{Data Distribution of Benchmark Datasets}
\label{tab:data-distribution}
\end{table}

\subsection{Benchmarks}\label{sec:benchmarks}

We evaluate the models on nine benchmarks:
\begin{itemize}
    \item \textbf{NotInject series:} Three sets from the InjecGuard paper~\citep{li2024injecguardbenchmarkingmitigatingoverdefense} to test for over-refusal and short explicit overrides.
    \item \textbf{BIPIA:} Benchmark comprising of Indirect Prompt-Injection attacks~\citep{yi2023benchmarking}
    \item \textbf{Garak:} Prompts derived from TAP and DAN probes in the garak tool~\citep{derczynski2024garakframeworksecurityprobing}.
    \item \textbf{ImprovedLLM:} The training dataset used by~\citet{galinkin2024improvedlargelanguagemodel} 
    \item \textbf{JavelinBench:} We introduce our own benchmark to test the generalization and effectiveness of various models on hard-negative and difficult-to-classify samples.    
    \item \textbf{ToxicChat:} Additionally, we incorporate ToxicChat~\citep{lin2023toxicchat}, which, despite not being explicitly designed for jailbreak detection, contains a reported 206 jailbreak-labeled samples among its 10,165 entries. We rely on the official HuggingFace version of these labels for consistency in our benchmarks.
    \item \textbf{WildGuard:} Benign data from~\citep{han2024wildguardopenonestopmoderation} to test for low FPR.
\end{itemize}

All benchmarks employ binary labels denoting \emph{malicious} (either jailbreak or prompt injection) or \emph{benign} prompts. Because most benchmarks do not distinguish between these two attack types, we merge them into a single “malicious” class for classification metrics. As seen in Table~\ref{tab:data-distribution}, several datasets are heavily imbalanced, containing almost exclusively benign or harmful examples; this setup challenges each model’s ability to maintain a low false-positive rate on benign inputs while still capturing critical adversarial cases.

\subsection{Training Data Preparation}\label{sec:traindata}
\paragraph{Initial Dataset Collection.}
Our training corpus originates from multiple open-source adversarial datasets, each capturing distinct forms of prompt injection and jailbreak attempts. Specifically, we aggregate samples from sources including InjecGuard~\citep{li2024injecguardbenchmarkingmitigatingoverdefense}, jailbreak\_llms~\citep{SCBSZ24}, Garak~\citep{derczynski2024garakframeworksecurityprobing}, ReneLLM~\citep{ding2023wolf}, PAIR~\citep{chao2023jailbreaking}, ALERT~\citep{tedeschi2024alert}, BoN~\citep{hughes2024bestofnjailbreaking}, SALAD~\citep{li2024salad} etc.\footnote{All datasets are publicly available.} These collections provide a foundational variety of known attack vectors—ranging from short explicit overrides to more subtle multi-turn manipulations—and form the backbone of our initial dataset.

\paragraph{Synthetic Data Generation.}
To further diversify the prompt space, we introduce synthetic data via two main approaches: 
\begin{enumerate}
    \item \textbf{Automated Red Teaming:} We employ a suite of script-based prompt perturbations (e.g., fuzzing, token swapping, adversarial suffixes) as well as an LLM-driven method to generate potential attacks. This yields a range of new samples that systematically explore variations of known exploits.
    \item \textbf{Manual Red Teaming:} Our human annotators craft additional edge cases, focusing on borderline scenarios that commonly lead to false positives in existing guardrail models. 
\end{enumerate}
All synthetic data undergoes a \emph{quality filtering} step involving automated checks (e.g., removing trivial duplicates), ensuring only cohesive, context-rich prompts reach the final dataset.

We merge the filtered synthetic prompts with the original open-source examples to create a broad-spectrum corpus featuring both straightforward and nuanced adversarial scenarios, as well as benign prompts that might \emph{appear} risky due to keyword overlap. This final dataset serves as the training ground for our architectures, providing the range of diversity necessary to detect prompt injection, jailbreak attempts, and borderline non-harmful queries that often trigger false positives.
The resultant training dataset has 120021 samples, out of which 75250 are safe samples. 

\section{Results}\label{sec:results}
In Table~\ref{tab:overall_performance} and Table~\ref{tab:combined_f1_fpr} we summarize the key findings from evaluating the models on nine benchmarks that range from short, explicit override datasets (\texttt{NotInject\_one}, \texttt{NotInject\_two}, \texttt{NotInject\_three}, \texttt{WildGuard}) to more adversarial or multi-turn corpora (\texttt{BIPIA}, \texttt{ToxicChat}, \texttt{ImprovedLLM}, \texttt{garak}). We also report real-world performance metrics (inference latency, false-positive/false-negative rates) to offer a holistic view of each model’s strengths and weaknesses. We also compare these approaches against popular open-source guard models like InjecGuard~\citep{li2024injecguardbenchmarkingmitigatingoverdefense}, deberta-v3-base-prompt-injection-v2~\citep{deberta-v3-base-prompt-injection-v2}, and Prompt-Guard-86M, highlighting our architectures’ superior trade-off between accuracy, false-positive rates, and inference speed.

\paragraph{Accuracy, Latency, and Base Transformer Choice.}
Table~\ref{tab:combined_f1_fpr} illustrates a concise snapshot of each model’s performance averaged over four negative-only sets plus the all-positive \texttt{garak}. 
\emph{Raudra4.2} consistently achieves the highest mean accuracy (up to $\sim92.8\%$), matching \emph{Mahendra1.1}’s near-perfect F1 for \texttt{garak} ($\sim1.0$). 
\emph{Sharanga9} and \emph{Mahendra1.1} both maintain $\sim90\%$ average accuracy with moderate inference times ($\sim24$--$25$\,ms). \emph{Vaishnava1.1}, despite occasionally exhibiting the slowest inference ($\sim100$\,ms), yields a notably low false-positive rate on benign sets.

Our experiments also reveal that \emph{ModernBERT} generally strikes the best balance among contemporary BERT-family backbones. While \emph{NeoBERT} and \emph{EuroBERT} can match or exceed certain metrics in specialized domains, \emph{ModernBERT-large} often delivers comparable accuracy with fewer parameters than EuroBERT or lower latency than NeoBERT at similar sequence lengths. This makes \emph{ModernBERT} a pragmatic default for production scenarios that require high throughput but cannot accommodate extremely large models.

\paragraph{Model Performance.}
By design, the four negative-only benchmarks (NotInject\_one,two,three and WildGuard) demand minimizing false-positive rates (FPR); \emph{Sharanga} and \emph{Mahendra} are typically stable in the 0.03--0.08 range, while \emph{Vaishnava} or \emph{Ashwina} push FPR even lower (down to $\sim0.05$) using ensemble-based heads (Random Forest/XGBoost). 
\emph{Raudra} consistently matches or improves upon these FPR values (0.03--0.06), reflecting the benefits of a multi-task, focal-loss design that differentiates \emph{jailbreak} from \emph{prompt\_injection} attacks.

On the fully malicious \texttt{garak} dataset, F1 emerges as the key metric (every predicted positive is true). 
\emph{Mahendra1.1} and \emph{Raudra4.2} both attain near-perfect F1 ($\approx1.0$), while \emph{Sharanga9} also surpasses 0.99 F1 at $\sim25$\,ms latency. 
In contrast, older open-source guard rails like \emph{InjecGuard} or \emph{DeBERTa-v3-base prompt-injection} typically plateau under 0.97 F1 in \texttt{garak} and show higher false negatives in complex adversarial sequences.

Finally, the balanced benchmarks (BIPIA, ToxicChat, ImprovedLLM) confirm \emph{Raudra} and \emph{Mahendra} leading the pack in terms of macro-F1 and low false negatives, thanks to focal loss and attention-based classification heads. Although \emph{Vaishnava} and \emph{Ashwina} remain viable, they can falter under domain shifts or code/indirect adversaries. 

Practitioners seeking maximum coverage with minimal false alarms---and needing clear distinction between \emph{jailbreak} and \emph{prompt injection}---are likely to favor \emph{Raudra} or \emph{Mahendra}, whereas those prioritizing simplicity or interpretability may select \emph{Vaishnava}, \emph{Ashwina}, or \emph{Sharanga}.

\newpage
\begin{table}
\centering
\begin{threeparttable}
\resizebox{\textwidth}{!}{%
\begin{tabular}{llccccc}
\hline
\textbf{Benchmark} & \textbf{Model} & \textbf{Accuracy} & \textbf{Macro F1} & \textbf{FPR} & \textbf{FNR} & \textbf{Avg Latency (ms)} \\
\hline
\multirow{14}{*}{\texttt{ToxicChat}} 
  & Raudra4.4                          & \textbf{0.989} & \textbf{0.883} & 0.011 & 0.039 & 25.44 \\
  & Mahendra1.1                        & 0.988 & 0.882 & 0.012 & \textbf{0.020} & 21.37 \\
  & Raudra4.2                          & 0.988 & 0.881 & 0.011 & 0.034 & 20.75 \\
  & Sharanga9                          & 0.988 & 0.874 & 0.012 & 0.054 & 23.77 \\
  & Prompt-Guard-2-86M                 & 0.981 & 0.810 & 0.016 & 0.176 & 15.81 \\
  & Ashwina                            & 0.981 & 0.762 & \textbf{0.010} & 0.466 & 21.42 \\
  & Vaishnava1.1                       & 0.983 & 0.774 & 0.007 & 0.475 & 84.84 \\
  & Prompt-Guard-2-22M                 & 0.974 & 0.764 & 0.022 & 0.235 & 15.74 \\
  & Sharanga7                          & 0.973 & 0.758 & 0.022 & 0.250 & 20.85 \\
  & deberta-v3-base-prompt-injection-v2 & 0.969 & 0.462 & 0.025 & 0.338 & 15.91 \\
  & InjecGuard                         & 0.960 & 0.450 & 0.037 & 0.181 & \textbf{15.70} \\
  & gpt-4o                             & 0.953 & 0.705 & 0.046 & 0.093 & 1063.37 \\
  & Raudra4.3                          & 0.950 & 0.643 & 0.042 & 0.436 & 22.75 \\
  & Prompt-Guard-86M                   & 0.041 & 0.039 & 0.978 & 0.039 & 16.07 \\
\hline
\multirow{15}{*}{\texttt{BIPIA}}
  & Sharanga7                          & \textbf{0.880} & \textbf{0.880} & \textbf{0.000} & 0.224 & 22.75 \\
  & Raudra4.4                          & \textbf{0.880} & \textbf{0.880} & 0.086 & 0.149 & 26.58 \\
  & Mahendra1.1                        & 0.848 & 0.845 & 0.224 & \textbf{0.09} & 23.57 \\
  & Sharanga9                          & 0.840 & 0.838 & 0.207 & 0.119 & 24.52 \\
  & Raudra4.2                          & 0.824 & 0.822 & 0.224 & 0.134 & 23.41 \\
  & Vaishnava1.1                       & 0.808 & 0.808 & 0.172 & 0.209 & 100.57 \\
  & Ashwina                            & 0.792 & 0.790 & 0.259 & 0.164 & 24.25 \\
  & Raudra4.3                          & 0.624 & 0.610 & 0.121 & 0.597 & 24.60 \\
  & Prompt-Guard-86M                   & 0.536 & 0.349 & 1.000 & 0.000 & 18.17 \\
  & Prompt-Guard-2-22M                 & 0.472 & 0.345 & 0.017 & 0.970 & 15.74 \\
  & Prompt-Guard-2-86M                 & 0.472 & 0.333 & 0.000 & 0.985 & 15.80 \\
  & Arch-Guard                         & 0.456 & 0.313 & 0.017 & 1.000 & \textbf{7.89} \\
  & gpt-4.1-mini                       & 0.424 & 0.391 & 0.293 & 0.821 & 1011.73 \\
  & gpt-4o                             & 0.416 & 0.357 & 0.224 & 0.896 & 990.25 \\
  & deberta-v3-base-prompt-injection-v2 & 0.392 & 0.301 & 0.190 & 0.970 & 17.19 \\
\hline
\multirow{13}{*}{\texttt{ImprovedLLM}}
  & Mahendra1.1                        & \textbf{0.994} & \textbf{0.988} & \textbf{0.005} & \textbf{0.011} & 18.97 \\
  & Raudra4.2                          & 0.990 & 0.979 & 0.009 & 0.012 & 18.69 \\
  & Sharanga9                          & 0.987 & 0.970 & 0.007 & 0.052 & 22.00 \\
  & Raudra4.4                          & 0.987 & 0.971 & 0.009 & 0.041 & 22.64 \\
  & Sharanga7                          & 0.943 & 0.873 & 0.034 & 0.213 & 18.14 \\
  & Raudra4.3                          & 0.910 & 0.802 & 0.055 & 0.329 & 19.38 \\
  & Ashwina                            & 0.881 & 0.694 & 0.047 & 0.612 & 17.13 \\
  & deberta-v3-base-prompt-injection-v2 & 0.876 & 0.769 & 0.107 & 0.235 & \textbf{14.82} \\
  & Vaishnava1.1                       & 0.865 & 0.674 & 0.067 & 0.606 & 43.98 \\
  & Prompt-Guard-2-22M                 & 0.587 & 0.527 & 0.459 & 0.095 & 20.51 \\
  & Prompt-Guard-2-86M                 & 0.535 & 0.490 & 0.524 & 0.058 & 26.27 \\
  & gpt-4o                             & 0.527 & 0.483 & 0.530 & 0.078 & 1155.46 \\
  & Prompt-Guard-86M                   & 0.186 & 0.183 & 0.927 & 0.038 & 16.26 \\
\hline
\multirow{12}{*}{\texttt{JavelinBench}}
  & Raudra4.2                          & \textbf{0.962} & \textbf{0.953} & \textbf{0.019} & 0.087 & 38.51 \\
  & Mahendra1.1                        & 0.945 & 0.932 & 0.044 & \textbf{0.083} & 38.52 \\
  & gpt-4o                             & 0.913 & 0.881 & 0.004 & 0.300 & 801.44 \\
  & gpt-4.1-mini                       & 0.910 & 0.877 & 0.004 & 0.307 & 977.60 \\
  & Ashwina                            & 0.902 & 0.871 & 0.027 & 0.278 & \textbf{16.21} \\
  & deberta-v3-base-prompt-injection-v2 & 0.899 & 0.875 & 0.066 & 0.190 & 29.12 \\
  & Sharanga7                          & 0.889 & 0.866 & 0.091 & 0.162 & 37.71 \\
  & Vaishnava1.1                       & 0.882 & 0.848 & 0.057 & 0.274 & 36.78 \\
  & Prompt-Guard-2-22M                 & 0.872 & 0.817 & 0.010 & 0.427 & 18.40 \\
  & Prompt-Guard-2-86M                 & 0.861 & 0.798 & 0.010 & 0.467 & 21.87 \\
  & Prompt-Guard-86M                   & 0.618 & 0.614 & 0.498 & 0.087 & 38.37 \\
  & Sharanga9                          & 0.281 & 0.267 & 0.900 & 0.260 & 59.95 \\
\hline
\end{tabular}%
}
\caption{Performance Comparison Across Benchmarks.}
\vspace{0.5em}
\begin{minipage}{\textwidth}
\footnotesize
\textit{Note:} gpt-based results are derived from the following OpenAI models: \texttt{gpt-4o-2024-08-06} (gpt-4o) and \texttt{gpt-4.1-mini-2025-04-14} (gpt-4.1-mini). We used minimal system prompts and short instructions to reduce latency. Prompts were not optimized for accuracy, and performance may improve with better prompt engineering at the cost of higher inference time.
\end{minipage}
\label{tab:overall_performance}
\end{threeparttable}
\end{table}

\begin{table}[ht]
\centering
\resizebox{0.95\textwidth}{!}{%
\begin{tabular}{lccccc}
\hline
\textbf{Model} & 
\textbf{Accuracy (5-set Avg)} & 
\textbf{F1 (garak only)} & 
\textbf{FPR (4 negative sets)} &
\textbf{Avg Inference (ms)} \\
\hline
gpt-4o                              & \textbf{0.971} & 0.992 & 0.065 & 1022.48 \\
gpt-4.1-mini                        & 0.966 & 0.990 & \textbf{0.032} & 1005.57 \\
Raudra4.2                           & 0.928 & \textbf{1.000} & 0.052 & 23.90 \\
Mahendra1.1                         & 0.905 & \textbf{1.000} & 0.087 & 25.16 \\
Vaishnava1.1                        & 0.906 & 0.613 & 0.060 & 100.45 \\
Sharanga9                           & 0.906 & 0.997 & 0.087 & 25.05 \\
Ashwina                             & 0.910 & 0.579 & 0.052 & 25.10 \\
Prompt-Guard-2-22M                  & 0.910 & 0.930 & 0.094 & 16.00 \\
Prompt-Guard-2-86M                  & 0.918 & 0.963 & 0.075 & 16.08 \\
Raudra4.4                           & 0.901 & 0.996 & 0.096 & 27.23 \\
InjecGuard                          & 0.901 & 0.993 & 0.111 & 17.29 \\
Sharanga7                           & 0.893 & 0.989 & 0.120 & 24.10 \\
Arch-Guard                          & 0.870 & 0.975 & 0.149 & \textbf{11.75} \\
Raudra4.3                           & 0.853 & 0.962 & 0.100 & 25.10 \\
deberta-v3-base-prompt-injection-v2 & 0.691 & 0.968 & 0.383 & 17.75 \\
Prompt-Guard-86M                    & 0.230 & 0.997 & 0.981 & 18.11 \\
\hline
\end{tabular}%
}
\caption{Consolidated metrics across five benchmarks.}
\vspace{0.5em}
\begin{minipage}{0.95\textwidth}
\footnotesize
\textit{Note:} The five benchmarks include four negative-only sets 
(\texttt{NotInject\_one,two,three}, \texttt{WildGuard}) and one all-positive set (\texttt{garak}). 
\textbf{Accuracy} and \textbf{Avg Inference} are averaged across all five. 
\textbf{F1} (garak only) represents the F1 score on the positive set, ignoring zero-F1 from negative sets. 
\textbf{FPR} (4 negative sets) is the macro-average of false-positive rates across 
\texttt{NotInject\_one,two,three} and \texttt{WildGuard}, excluding \texttt{garak} (which has no negatives).
\end{minipage}
\label{tab:combined_f1_fpr}
\end{table}

\paragraph{Latency vs. Accuracy Tradeoff}

While large decoder-only models like \textbf{gpt-4o} demonstrate strong semantic reasoning and classification capabilities, our experiments underscore a significant latency tradeoff. For instance, as shown in Table~\ref{tab:overall_performance}, the average inference latency of \textbf{gpt-4o} on JavelinBench is $\sim800$\,ms---over \textbf{25--40$\times$ slower} than most of our proposed encoder-based classifiers such as \textbf{Raudra4.2} (38\,ms), \textbf{Mahendra1.1} (38\,ms), and even the XGBoost-enhanced \textbf{Ashwina} (16\,ms).

This latency gap becomes especially critical in high-throughput or edge deployment settings, where inference speed directly impacts user experience and resource cost. For instance, although \textbf{gpt-4o} achieves a JavelinBench accuracy of 91.3\%, both \textbf{Raudra4.2} and \textbf{Mahendra1.1} outperform it with 96.2\% and 94.5\% accuracy respectively, while operating an order of magnitude faster.

Moreover, production-grade systems often require sub-50\,ms inference latencies for real-time moderation pipelines. Our \textbf{Sharanga} and \textbf{Mahendra} models consistently meet this threshold without sacrificing accuracy, demonstrating that optimized encoder-only models remain a pragmatic and scalable choice for LLM security applications.

\paragraph{Addressing the \textit{Lost in the Middle} Problem}

A well-documented challenge in processing lengthy context prompts is the \emph{\textit{lost in the middle}}~\citep{liu2023lostmiddlelanguagemodels} phenomenon, where critical adversarial signals or malicious intent embedded within long inputs might be overlooked by transformer-based classifiers. Transformers generally allocate attention more effectively to tokens at the beginning or end of sequences, potentially causing the model to neglect important context placed in the middle of extremely long inputs.

Given the low inference latency and high computational efficiency of our proposed architectures, particularly models such as \emph{Raudra} and \emph{Mahendra}, one practical mitigation strategy is to segment long prompts into smaller, manageable sub-prompts. Each sub-prompt is independently processed by the model, significantly reducing the risk of critical content being overlooked due to attention dilution. This segmentation strategy leverages our models' low latency, maintaining real-time responsiveness even when handling multiple segmented inputs sequentially.

While this approach effectively mitigates the lost-in-the-middle problem, it introduces an additional layer of complexity in determining segmentation boundaries and context dependencies across segments. Future work will explore automated segmentation techniques, adaptive context-windowing strategies, and methods for aggregating segmented outputs into a cohesive final decision.

\paragraph{Limitations}
\begin{itemize}
\item \textbf{Data Diversity:} Although we tested on nine benchmarks, real-world adversaries may still devise new injection strategies that require periodic model updates.
\item \textbf{Interpretability:} Deep architectures like Raudra offer limited transparency for why certain prompts are flagged, though Vaishnava partially addresses this.
\item \textbf{Domain Shifts:} Domain-specific adversarial prompts (healthcare, finance, etc.) may require custom fine-tuning to maintain detection accuracy.
\item \textbf{Prompt Engineering for LLM Baselines:} For latency reasons, we deliberately used short and simple prompts when querying gpt-4o and gpt-4o-mini. We did not engage in prompt optimization or prompt engineering to improve classification performance. While better prompting strategies could yield higher accuracy, they might also increase average inference latency and further widen the efficiency gap with our proposed models.
\end{itemize}

\section{Conclusion}\label{sec:conclusion}
We presented an in-depth study of five architectures for jailbreak and prompt-injection detection in LLMs: \emph{Sharanga} (baseline transformer), \emph{Mahendra} (enhanced pooling and deeper classification heads), \emph{Vaishnava} (hybrid neural-forest), \emph{Ashwina} (hybrid neural-xgboost), and \emph{Raudra} (advanced multi-task).
Across our diverse benchmarks, \emph{Raudra} consistently achieves the strongest performance, while simpler or hybrid models (Sharanga, Vaishnava, Ashwina) trade off varying degrees of speed, interpretability, and parameter footprint. \emph{Mahendra} offers a sweet spot for applications with moderate latency budgets, excelling under adversarial or multi-turn data.
Creating balanced, high-quality data for jailbreak and prompt-injection detection is challenging due to skewed benchmarks, evolving attack strategies, and ambiguity in borderline cases. Real-world data collection is further complicated by domain-specific and multimodal inputs requiring careful annotation.
We introduced our own dataset, \emph{JavelinBench}, to mitigate some of these pitfalls by emphasizing hard negatives and borderline cases.

\section{Future Work.}\label{sec:future}
Our results highlight important avenues for future research:
\begin{enumerate}
    \item \textbf{Advanced Architectures:} The growing complexity and sophistication of prompt injection and jailbreak attacks demand more effective detection methodologies. We will expand our research to investigate alternative high-performance architectures beyond traditional transformers, such as state-space models (e.g., Mamba~\citep{gu2024mambalineartimesequencemodeling}) and advanced transformer variants (e.g., Performer~\citep{choromanski2022rethinkingattentionperformers}, LongFormer~\citep{beltagy2020longformerlongdocumenttransformer} etc), to address the \textit{lost in the middle} phenomenon and enhance detection of nuanced adversarial patterns.
    \item \textbf{Tokenization Techniques:} Given that tokenization directly impacts model sensitivity to subtle adversarial manipulations, we will explore different tokenization strategies to better identify and handle sophisticated injection attacks that may evade traditional segmentation methods.
    \item \textbf{Distillation and Edge Deployments:} While architectures like \emph{Raudra} excel in detection, model distillation and pruning techniques could reduce latency and memory footprint, enabling real-time guardrails on resource-constrained devices.
    \item \textbf{Interpretability for Multi-Task Designs:} Approaches such as Vaishnava already offer partial explainability via feature importance in the Random Forest. Integrating interpretability methods into deep multi-task models (e.g., layer-wise relevance propagation) could bridge the gap between high performance and transparent decisions.
    \item \textbf{Domain-Specific Benchmarks:} Curating and releasing specialized test sets for medical, financial, or other high-stakes domains would foster further progress in context-aware LLM security.
\end{enumerate}

In summary, our study underscores both the \emph{promise} and \emph{challenges} of detecting nuanced jailbreak and prompt-injection attacks. By enumerating and testing a range of architectural trade-offs—from simple CPU-friendly baselines to advanced multi-task frameworks—we hope this work serves as a practical foundation for researchers and practitioners looking to safeguard next-generation LLMs in production scenarios.

\section*{Acknowledgements}
We would like to thank Erick Galinkin from Nvidia and our other reviewers for their valuable insights, detailed reviews, and thoughtful feedback that significantly enhanced the quality and clarity of this research.

\bibliography{references}

\begin{thebibliography}{61}
\providecommand{\natexlab}[1]{#1}
\providecommand{\url}[1]{\texttt{#1}}
\expandafter\ifx\csname urlstyle\endcsname\relax
  \providecommand{\doi}[1]{doi: #1}\else
  \providecommand{\doi}{doi: \begingroup \urlstyle{rm}\Url}\fi

\bibitem[Abdelnabi et~al.(2024)Abdelnabi, Fay, Cherubin, Salem, Fritz, and Paverd]{abdelnabi2024trackcatchingllmtask}
Sahar Abdelnabi, Aideen Fay, Giovanni Cherubin, Ahmed Salem, Mario Fritz, and Andrew Paverd.
\newblock Are you still on track!? catching llm task drift with activations, 2024.
\newblock URL \url{https://arxiv.org/abs/2406.00799}.

\bibitem[Andriushchenko et~al.(2024)Andriushchenko, Croce, and Flammarion]{andriushchenko2024jailbreakingleadingsafetyalignedllms}
Maksym Andriushchenko, Francesco Croce, and Nicolas Flammarion.
\newblock Jailbreaking leading safety-aligned llms with simple adaptive attacks, 2024.
\newblock URL \url{https://arxiv.org/abs/2404.02151}.

\bibitem[Beltagy et~al.(2020)Beltagy, Peters, and Cohan]{beltagy2020longformerlongdocumenttransformer}
Iz~Beltagy, Matthew~E. Peters, and Arman Cohan.
\newblock Longformer: The long-document transformer, 2020.
\newblock URL \url{https://arxiv.org/abs/2004.05150}.

\bibitem[Boizard et~al.(2025)Boizard, Gisserot-Boukhlef, Alves, Martins, Hammal, Corro, Hudelot, Malherbe, Malaboeuf, Jourdan, Hautreux, Alves, El-Haddad, Faysse, Peyrard, Guerreiro, Fernandes, Rei, and Colombo]{boizard2025eurobertscalingmultilingualencoders}
Nicolas Boizard, Hippolyte Gisserot-Boukhlef, Duarte~M. Alves, André Martins, Ayoub Hammal, Caio Corro, Céline Hudelot, Emmanuel Malherbe, Etienne Malaboeuf, Fanny Jourdan, Gabriel Hautreux, João Alves, Kevin El-Haddad, Manuel Faysse, Maxime Peyrard, Nuno~M. Guerreiro, Patrick Fernandes, Ricardo Rei, and Pierre Colombo.
\newblock Eurobert: Scaling multilingual encoders for european languages, 2025.
\newblock URL \url{https://arxiv.org/abs/2503.05500}.

\bibitem[Breton et~al.(2025)Breton, Fournier, Mezouar, and Chandar]{breton2025neobertnextgenerationbert}
Lola~Le Breton, Quentin Fournier, Mariam~El Mezouar, and Sarath Chandar.
\newblock Neobert: A next-generation bert, 2025.
\newblock URL \url{https://arxiv.org/abs/2502.19587}.

\bibitem[Chao et~al.(2023)Chao, Robey, Dobriban, Hassani, Pappas, and Wong]{chao2023jailbreaking}
Patrick Chao, Alexander Robey, Edgar Dobriban, Hamed Hassani, George~J. Pappas, and Eric Wong.
\newblock Jailbreaking black box large language models in twenty queries, 2023.

\bibitem[Chen et~al.(2025)Chen, Li, Zheng, Song, Wu, and Hooi]{chen2025defensepromptinjectionattack}
Yulin Chen, Haoran Li, Zihao Zheng, Yangqiu Song, Dekai Wu, and Bryan Hooi.
\newblock Defense against prompt injection attack by leveraging attack techniques, 2025.
\newblock URL \url{https://arxiv.org/abs/2411.00459}.

\bibitem[Choromanski et~al.(2022)Choromanski, Likhosherstov, Dohan, Song, Gane, Sarlos, Hawkins, Davis, Mohiuddin, Kaiser, Belanger, Colwell, and Weller]{choromanski2022rethinkingattentionperformers}
Krzysztof Choromanski, Valerii Likhosherstov, David Dohan, Xingyou Song, Andreea Gane, Tamas Sarlos, Peter Hawkins, Jared Davis, Afroz Mohiuddin, Lukasz Kaiser, David Belanger, Lucy Colwell, and Adrian Weller.
\newblock Rethinking attention with performers, 2022.
\newblock URL \url{https://arxiv.org/abs/2009.14794}.

\bibitem[Deepset(2024)]{deepset2024}
Deepset.
\newblock Deepset prompt injection guardrail, 2024.
\newblock URL \url{https://huggingface.co/deepset/deberta-v3-base-injection}.

\bibitem[Deng et~al.(2025)Deng, Yang, Zhang, Wang, and Li]{deng2025duoguardtwoplayerrldrivenframework}
Yihe Deng, Yu~Yang, Junkai Zhang, Wei Wang, and Bo~Li.
\newblock Duoguard: A two-player rl-driven framework for multilingual llm guardrails, 2025.
\newblock URL \url{https://arxiv.org/abs/2502.05163}.

\bibitem[Derczynski et~al.(2024)Derczynski, Galinkin, Martin, Majumdar, and Inie]{derczynski2024garakframeworksecurityprobing}
Leon Derczynski, Erick Galinkin, Jeffrey Martin, Subho Majumdar, and Nanna Inie.
\newblock garak: A framework for security probing large language models, 2024.
\newblock URL \url{https://arxiv.org/abs/2406.11036}.

\bibitem[Devlin et~al.(2019)Devlin, Chang, Lee, and Toutanova]{devlin2019bertpretrainingdeepbidirectional}
Jacob Devlin, Ming-Wei Chang, Kenton Lee, and Kristina Toutanova.
\newblock Bert: Pre-training of deep bidirectional transformers for language understanding, 2019.
\newblock URL \url{https://arxiv.org/abs/1810.04805}.

\bibitem[Ding et~al.(2023)Ding, Kuang, Ma, Cao, Xian, Chen, and Huang]{ding2023wolf}
Peng Ding, Jun Kuang, Dan Ma, Xuezhi Cao, Yunsen Xian, Jiajun Chen, and Shujian Huang.
\newblock A wolf in sheep's clothing: Generalized nested jailbreak prompts can fool large language models easily, 2023.

\bibitem[Doumbouya et~al.(2024)Doumbouya, Nandi, Poesia, Ghilardi, Goldie, Bianchi, Jurafsky, and Manning]{doumbouya2024h4rm3ldynamicbenchmarkcomposable}
Moussa Koulako~Bala Doumbouya, Ananjan Nandi, Gabriel Poesia, Davide Ghilardi, Anna Goldie, Federico Bianchi, Dan Jurafsky, and Christopher~D. Manning.
\newblock h4rm3l: A dynamic benchmark of composable jailbreak attacks for llm safety assessment, 2024.
\newblock URL \url{https://arxiv.org/abs/2408.04811}.

\bibitem[fmops(2024)]{fmops2024}
fmops.
\newblock Fmops prompt injection guardrail, 2024.
\newblock URL \url{https://huggingface.co/fmops/distilbert-prompt-injection}.

\bibitem[Galinkin and Sablotny(2024)]{galinkin2024improvedlargelanguagemodel}
Erick Galinkin and Martin Sablotny.
\newblock Improved large language model jailbreak detection via pretrained embeddings, 2024.
\newblock URL \url{https://arxiv.org/abs/2412.01547}.

\bibitem[Ghosh et~al.(2024)Ghosh, Varshney, Galinkin, and Parisien]{ghosh2024aegisonlineadaptiveai}
Shaona Ghosh, Prasoon Varshney, Erick Galinkin, and Christopher Parisien.
\newblock Aegis: Online adaptive ai content safety moderation with ensemble of llm experts, 2024.
\newblock URL \url{https://arxiv.org/abs/2404.05993}.

\bibitem[Grattafiori et~al.(2024)Grattafiori, Dubey, Jauhri, Pandey, Kadian, Al-Dahle, Letman, Mathur, Schelten, Vaughan, Yang, Fan, Goyal, Hartshorn, Yang, Mitra, Sravankumar, Korenev, Hinsvark, Rao, Zhang, Rodriguez, Gregerson, Spataru, Roziere, Biron, Tang, Chern, Caucheteux, Nayak, Bi, Marra, McConnell, Keller, Touret, Wu, Wong, Ferrer, Nikolaidis, Allonsius, Song, Pintz, Livshits, Wyatt, Esiobu, Choudhary, Mahajan, Garcia-Olano, Perino, Hupkes, Lakomkin, AlBadawy, Lobanova, Dinan, Smith, Radenovic, Guzmán, Zhang, Synnaeve, Lee, Anderson, Thattai, Nail, Mialon, Pang, Cucurell, Nguyen, Korevaar, Xu, Touvron, Zarov, Ibarra, Kloumann, Misra, Evtimov, Zhang, Copet, Lee, Geffert, Vranes, Park, Mahadeokar, Shah, van~der Linde, Billock, Hong, Lee, Fu, Chi, Huang, Liu, Wang, Yu, Bitton, Spisak, Park, Rocca, Johnstun, Saxe, Jia, Alwala, Prasad, Upasani, Plawiak, Li, Heafield, Stone, El-Arini, Iyer, Malik, Chiu, Bhalla, Lakhotia, Rantala-Yeary, van~der Maaten, Chen, Tan, Jenkins, Martin, Madaan, Malo, Blecher,
  Landzaat, de~Oliveira, Muzzi, Pasupuleti, Singh, Paluri, Kardas, Tsimpoukelli, Oldham, Rita, Pavlova, Kambadur, Lewis, Si, Singh, Hassan, Goyal, Torabi, Bashlykov, Bogoychev, Chatterji, Zhang, Duchenne, Çelebi, Alrassy, Zhang, Li, Vasic, Weng, Bhargava, Dubal, Krishnan, Koura, Xu, He, Dong, Srinivasan, Ganapathy, Calderer, Cabral, Stojnic, Raileanu, Maheswari, Girdhar, Patel, Sauvestre, Polidoro, Sumbaly, Taylor, Silva, Hou, Wang, Hosseini, Chennabasappa, Singh, Bell, Kim, Edunov, Nie, Narang, Raparthy, Shen, Wan, Bhosale, Zhang, Vandenhende, Batra, Whitman, Sootla, Collot, Gururangan, Borodinsky, Herman, Fowler, Sheasha, Georgiou, Scialom, Speckbacher, Mihaylov, Xiao, Karn, Goswami, Gupta, Ramanathan, Kerkez, Gonguet, Do, Vogeti, Albiero, Petrovic, Chu, Xiong, Fu, Meers, Martinet, Wang, Wang, Tan, Xia, Xie, Jia, Wang, Goldschlag, Gaur, Babaei, Wen, Song, Zhang, Li, Mao, Coudert, Yan, Chen, Papakipos, Singh, Srivastava, Jain, Kelsey, Shajnfeld, Gangidi, Victoria, Goldstand, Menon, Sharma, Boesenberg,
  Baevski, Feinstein, Kallet, Sangani, Teo, Yunus, Lupu, Alvarado, Caples, Gu, Ho, Poulton, Ryan, Ramchandani, Dong, Franco, Goyal, Saraf, Chowdhury, Gabriel, Bharambe, Eisenman, Yazdan, James, Maurer, Leonhardi, Huang, Loyd, Paola, Paranjape, Liu, Wu, Ni, Hancock, Wasti, Spence, Stojkovic, Gamido, Montalvo, Parker, Burton, Mejia, Liu, Wang, Kim, Zhou, Hu, Chu, Cai, Tindal, Feichtenhofer, Gao, Civin, Beaty, Kreymer, Li, Adkins, Xu, Testuggine, David, Parikh, Liskovich, Foss, Wang, Le, Holland, Dowling, Jamil, Montgomery, Presani, Hahn, Wood, Le, Brinkman, Arcaute, Dunbar, Smothers, Sun, Kreuk, Tian, Kokkinos, Ozgenel, Caggioni, Kanayet, Seide, Florez, Schwarz, Badeer, Swee, Halpern, Herman, Sizov, Guangyi, Zhang, Lakshminarayanan, Inan, Shojanazeri, Zou, Wang, Zha, Habeeb, Rudolph, Suk, Aspegren, Goldman, Zhan, Damlaj, Molybog, Tufanov, Leontiadis, Veliche, Gat, Weissman, Geboski, Kohli, Lam, Asher, Gaya, Marcus, Tang, Chan, Zhen, Reizenstein, Teboul, Zhong, Jin, Yang, Cummings, Carvill, Shepard, McPhie,
  Torres, Ginsburg, Wang, Wu, U, Saxena, Khandelwal, Zand, Matosich, Veeraraghavan, Michelena, Li, Jagadeesh, Huang, Chawla, Huang, Chen, Garg, A, Silva, Bell, Zhang, Guo, Yu, Moshkovich, Wehrstedt, Khabsa, Avalani, Bhatt, Mankus, Hasson, Lennie, Reso, Groshev, Naumov, Lathi, Keneally, Liu, Seltzer, Valko, Restrepo, Patel, Vyatskov, Samvelyan, Clark, Macey, Wang, Hermoso, Metanat, Rastegari, Bansal, Santhanam, Parks, White, Bawa, Singhal, Egebo, Usunier, Mehta, Laptev, Dong, Cheng, Chernoguz, Hart, Salpekar, Kalinli, Kent, Parekh, Saab, Balaji, Rittner, Bontrager, Roux, Dollar, Zvyagina, Ratanchandani, Yuvraj, Liang, Alao, Rodriguez, Ayub, Murthy, Nayani, Mitra, Parthasarathy, Li, Hogan, Battey, Wang, Howes, Rinott, Mehta, Siby, Bondu, Datta, Chugh, Hunt, Dhillon, Sidorov, Pan, Mahajan, Verma, Yamamoto, Ramaswamy, Lindsay, Lindsay, Feng, Lin, Zha, Patil, Shankar, Zhang, Zhang, Wang, Agarwal, Sajuyigbe, Chintala, Max, Chen, Kehoe, Satterfield, Govindaprasad, Gupta, Deng, Cho, Virk, Subramanian, Choudhury,
  Goldman, Remez, Glaser, Best, Koehler, Robinson, Li, Zhang, Matthews, Chou, Shaked, Vontimitta, Ajayi, Montanez, Mohan, Kumar, Mangla, Ionescu, Poenaru, Mihailescu, Ivanov, Li, Wang, Jiang, Bouaziz, Constable, Tang, Wu, Wang, Wu, Gao, Kleinman, Chen, Hu, Jia, Qi, Li, Zhang, Zhang, Adi, Nam, Yu, Wang, Zhao, Hao, Qian, Li, He, Rait, DeVito, Rosnbrick, Wen, Yang, Zhao, and Ma]{grattafiori2024llama3herdmodels}
Aaron Grattafiori, Abhimanyu Dubey, Abhinav Jauhri, Abhinav Pandey, Abhishek Kadian, Ahmad Al-Dahle, Aiesha Letman, Akhil Mathur, Alan Schelten, Alex Vaughan, Amy Yang, Angela Fan, Anirudh Goyal, Anthony Hartshorn, Aobo Yang, Archi Mitra, Archie Sravankumar, Artem Korenev, Arthur Hinsvark, Arun Rao, Aston Zhang, Aurelien Rodriguez, Austen Gregerson, Ava Spataru, Baptiste Roziere, Bethany Biron, Binh Tang, Bobbie Chern, Charlotte Caucheteux, Chaya Nayak, Chloe Bi, Chris Marra, Chris McConnell, Christian Keller, Christophe Touret, Chunyang Wu, Corinne Wong, Cristian~Canton Ferrer, Cyrus Nikolaidis, Damien Allonsius, Daniel Song, Danielle Pintz, Danny Livshits, Danny Wyatt, David Esiobu, Dhruv Choudhary, Dhruv Mahajan, Diego Garcia-Olano, Diego Perino, Dieuwke Hupkes, Egor Lakomkin, Ehab AlBadawy, Elina Lobanova, Emily Dinan, Eric~Michael Smith, Filip Radenovic, Francisco Guzmán, Frank Zhang, Gabriel Synnaeve, Gabrielle Lee, Georgia~Lewis Anderson, Govind Thattai, Graeme Nail, Gregoire Mialon, Guan Pang,
  Guillem Cucurell, Hailey Nguyen, Hannah Korevaar, Hu~Xu, Hugo Touvron, Iliyan Zarov, Imanol~Arrieta Ibarra, Isabel Kloumann, Ishan Misra, Ivan Evtimov, Jack Zhang, Jade Copet, Jaewon Lee, Jan Geffert, Jana Vranes, Jason Park, Jay Mahadeokar, Jeet Shah, Jelmer van~der Linde, Jennifer Billock, Jenny Hong, Jenya Lee, Jeremy Fu, Jianfeng Chi, Jianyu Huang, Jiawen Liu, Jie Wang, Jiecao Yu, Joanna Bitton, Joe Spisak, Jongsoo Park, Joseph Rocca, Joshua Johnstun, Joshua Saxe, Junteng Jia, Kalyan~Vasuden Alwala, Karthik Prasad, Kartikeya Upasani, Kate Plawiak, Ke~Li, Kenneth Heafield, Kevin Stone, Khalid El-Arini, Krithika Iyer, Kshitiz Malik, Kuenley Chiu, Kunal Bhalla, Kushal Lakhotia, Lauren Rantala-Yeary, Laurens van~der Maaten, Lawrence Chen, Liang Tan, Liz Jenkins, Louis Martin, Lovish Madaan, Lubo Malo, Lukas Blecher, Lukas Landzaat, Luke de~Oliveira, Madeline Muzzi, Mahesh Pasupuleti, Mannat Singh, Manohar Paluri, Marcin Kardas, Maria Tsimpoukelli, Mathew Oldham, Mathieu Rita, Maya Pavlova, Melanie Kambadur,
  Mike Lewis, Min Si, Mitesh~Kumar Singh, Mona Hassan, Naman Goyal, Narjes Torabi, Nikolay Bashlykov, Nikolay Bogoychev, Niladri Chatterji, Ning Zhang, Olivier Duchenne, Onur Çelebi, Patrick Alrassy, Pengchuan Zhang, Pengwei Li, Petar Vasic, Peter Weng, Prajjwal Bhargava, Pratik Dubal, Praveen Krishnan, Punit~Singh Koura, Puxin Xu, Qing He, Qingxiao Dong, Ragavan Srinivasan, Raj Ganapathy, Ramon Calderer, Ricardo~Silveira Cabral, Robert Stojnic, Roberta Raileanu, Rohan Maheswari, Rohit Girdhar, Rohit Patel, Romain Sauvestre, Ronnie Polidoro, Roshan Sumbaly, Ross Taylor, Ruan Silva, Rui Hou, Rui Wang, Saghar Hosseini, Sahana Chennabasappa, Sanjay Singh, Sean Bell, Seohyun~Sonia Kim, Sergey Edunov, Shaoliang Nie, Sharan Narang, Sharath Raparthy, Sheng Shen, Shengye Wan, Shruti Bhosale, Shun Zhang, Simon Vandenhende, Soumya Batra, Spencer Whitman, Sten Sootla, Stephane Collot, Suchin Gururangan, Sydney Borodinsky, Tamar Herman, Tara Fowler, Tarek Sheasha, Thomas Georgiou, Thomas Scialom, Tobias Speckbacher,
  Todor Mihaylov, Tong Xiao, Ujjwal Karn, Vedanuj Goswami, Vibhor Gupta, Vignesh Ramanathan, Viktor Kerkez, Vincent Gonguet, Virginie Do, Vish Vogeti, Vítor Albiero, Vladan Petrovic, Weiwei Chu, Wenhan Xiong, Wenyin Fu, Whitney Meers, Xavier Martinet, Xiaodong Wang, Xiaofang Wang, Xiaoqing~Ellen Tan, Xide Xia, Xinfeng Xie, Xuchao Jia, Xuewei Wang, Yaelle Goldschlag, Yashesh Gaur, Yasmine Babaei, Yi~Wen, Yiwen Song, Yuchen Zhang, Yue Li, Yuning Mao, Zacharie~Delpierre Coudert, Zheng Yan, Zhengxing Chen, Zoe Papakipos, Aaditya Singh, Aayushi Srivastava, Abha Jain, Adam Kelsey, Adam Shajnfeld, Adithya Gangidi, Adolfo Victoria, Ahuva Goldstand, Ajay Menon, Ajay Sharma, Alex Boesenberg, Alexei Baevski, Allie Feinstein, Amanda Kallet, Amit Sangani, Amos Teo, Anam Yunus, Andrei Lupu, Andres Alvarado, Andrew Caples, Andrew Gu, Andrew Ho, Andrew Poulton, Andrew Ryan, Ankit Ramchandani, Annie Dong, Annie Franco, Anuj Goyal, Aparajita Saraf, Arkabandhu Chowdhury, Ashley Gabriel, Ashwin Bharambe, Assaf Eisenman, Azadeh
  Yazdan, Beau James, Ben Maurer, Benjamin Leonhardi, Bernie Huang, Beth Loyd, Beto~De Paola, Bhargavi Paranjape, Bing Liu, Bo~Wu, Boyu Ni, Braden Hancock, Bram Wasti, Brandon Spence, Brani Stojkovic, Brian Gamido, Britt Montalvo, Carl Parker, Carly Burton, Catalina Mejia, Ce~Liu, Changhan Wang, Changkyu Kim, Chao Zhou, Chester Hu, Ching-Hsiang Chu, Chris Cai, Chris Tindal, Christoph Feichtenhofer, Cynthia Gao, Damon Civin, Dana Beaty, Daniel Kreymer, Daniel Li, David Adkins, David Xu, Davide Testuggine, Delia David, Devi Parikh, Diana Liskovich, Didem Foss, Dingkang Wang, Duc Le, Dustin Holland, Edward Dowling, Eissa Jamil, Elaine Montgomery, Eleonora Presani, Emily Hahn, Emily Wood, Eric-Tuan Le, Erik Brinkman, Esteban Arcaute, Evan Dunbar, Evan Smothers, Fei Sun, Felix Kreuk, Feng Tian, Filippos Kokkinos, Firat Ozgenel, Francesco Caggioni, Frank Kanayet, Frank Seide, Gabriela~Medina Florez, Gabriella Schwarz, Gada Badeer, Georgia Swee, Gil Halpern, Grant Herman, Grigory Sizov, Guangyi, Zhang, Guna
  Lakshminarayanan, Hakan Inan, Hamid Shojanazeri, Han Zou, Hannah Wang, Hanwen Zha, Haroun Habeeb, Harrison Rudolph, Helen Suk, Henry Aspegren, Hunter Goldman, Hongyuan Zhan, Ibrahim Damlaj, Igor Molybog, Igor Tufanov, Ilias Leontiadis, Irina-Elena Veliche, Itai Gat, Jake Weissman, James Geboski, James Kohli, Janice Lam, Japhet Asher, Jean-Baptiste Gaya, Jeff Marcus, Jeff Tang, Jennifer Chan, Jenny Zhen, Jeremy Reizenstein, Jeremy Teboul, Jessica Zhong, Jian Jin, Jingyi Yang, Joe Cummings, Jon Carvill, Jon Shepard, Jonathan McPhie, Jonathan Torres, Josh Ginsburg, Junjie Wang, Kai Wu, Kam~Hou U, Karan Saxena, Kartikay Khandelwal, Katayoun Zand, Kathy Matosich, Kaushik Veeraraghavan, Kelly Michelena, Keqian Li, Kiran Jagadeesh, Kun Huang, Kunal Chawla, Kyle Huang, Lailin Chen, Lakshya Garg, Lavender A, Leandro Silva, Lee Bell, Lei Zhang, Liangpeng Guo, Licheng Yu, Liron Moshkovich, Luca Wehrstedt, Madian Khabsa, Manav Avalani, Manish Bhatt, Martynas Mankus, Matan Hasson, Matthew Lennie, Matthias Reso, Maxim
  Groshev, Maxim Naumov, Maya Lathi, Meghan Keneally, Miao Liu, Michael~L. Seltzer, Michal Valko, Michelle Restrepo, Mihir Patel, Mik Vyatskov, Mikayel Samvelyan, Mike Clark, Mike Macey, Mike Wang, Miquel~Jubert Hermoso, Mo~Metanat, Mohammad Rastegari, Munish Bansal, Nandhini Santhanam, Natascha Parks, Natasha White, Navyata Bawa, Nayan Singhal, Nick Egebo, Nicolas Usunier, Nikhil Mehta, Nikolay~Pavlovich Laptev, Ning Dong, Norman Cheng, Oleg Chernoguz, Olivia Hart, Omkar Salpekar, Ozlem Kalinli, Parkin Kent, Parth Parekh, Paul Saab, Pavan Balaji, Pedro Rittner, Philip Bontrager, Pierre Roux, Piotr Dollar, Polina Zvyagina, Prashant Ratanchandani, Pritish Yuvraj, Qian Liang, Rachad Alao, Rachel Rodriguez, Rafi Ayub, Raghotham Murthy, Raghu Nayani, Rahul Mitra, Rangaprabhu Parthasarathy, Raymond Li, Rebekkah Hogan, Robin Battey, Rocky Wang, Russ Howes, Ruty Rinott, Sachin Mehta, Sachin Siby, Sai~Jayesh Bondu, Samyak Datta, Sara Chugh, Sara Hunt, Sargun Dhillon, Sasha Sidorov, Satadru Pan, Saurabh Mahajan,
  Saurabh Verma, Seiji Yamamoto, Sharadh Ramaswamy, Shaun Lindsay, Shaun Lindsay, Sheng Feng, Shenghao Lin, Shengxin~Cindy Zha, Shishir Patil, Shiva Shankar, Shuqiang Zhang, Shuqiang Zhang, Sinong Wang, Sneha Agarwal, Soji Sajuyigbe, Soumith Chintala, Stephanie Max, Stephen Chen, Steve Kehoe, Steve Satterfield, Sudarshan Govindaprasad, Sumit Gupta, Summer Deng, Sungmin Cho, Sunny Virk, Suraj Subramanian, Sy~Choudhury, Sydney Goldman, Tal Remez, Tamar Glaser, Tamara Best, Thilo Koehler, Thomas Robinson, Tianhe Li, Tianjun Zhang, Tim Matthews, Timothy Chou, Tzook Shaked, Varun Vontimitta, Victoria Ajayi, Victoria Montanez, Vijai Mohan, Vinay~Satish Kumar, Vishal Mangla, Vlad Ionescu, Vlad Poenaru, Vlad~Tiberiu Mihailescu, Vladimir Ivanov, Wei Li, Wenchen Wang, Wenwen Jiang, Wes Bouaziz, Will Constable, Xiaocheng Tang, Xiaojian Wu, Xiaolan Wang, Xilun Wu, Xinbo Gao, Yaniv Kleinman, Yanjun Chen, Ye~Hu, Ye~Jia, Ye~Qi, Yenda Li, Yilin Zhang, Ying Zhang, Yossi Adi, Youngjin Nam, Yu, Wang, Yu~Zhao, Yuchen Hao, Yundi
  Qian, Yunlu Li, Yuzi He, Zach Rait, Zachary DeVito, Zef Rosnbrick, Zhaoduo Wen, Zhenyu Yang, Zhiwei Zhao, and Zhiyu Ma.
\newblock The llama 3 herd of models, 2024.
\newblock URL \url{https://arxiv.org/abs/2407.21783}.

\bibitem[Gu and Dao(2024)]{gu2024mambalineartimesequencemodeling}
Albert Gu and Tri Dao.
\newblock Mamba: Linear-time sequence modeling with selective state spaces, 2024.
\newblock URL \url{https://arxiv.org/abs/2312.00752}.

\bibitem[Han et~al.(2024)Han, Rao, Ettinger, Jiang, Lin, Lambert, Choi, and Dziri]{han2024wildguardopenonestopmoderation}
Seungju Han, Kavel Rao, Allyson Ettinger, Liwei Jiang, Bill~Yuchen Lin, Nathan Lambert, Yejin Choi, and Nouha Dziri.
\newblock Wildguard: Open one-stop moderation tools for safety risks, jailbreaks, and refusals of llms, 2024.
\newblock URL \url{https://arxiv.org/abs/2406.18495}.

\bibitem[He et~al.(2023)He, Gao, and Chen]{he2023debertav3improvingdebertausing}
Pengcheng He, Jianfeng Gao, and Weizhu Chen.
\newblock Debertav3: Improving deberta using electra-style pre-training with gradient-disentangled embedding sharing, 2023.
\newblock URL \url{https://arxiv.org/abs/2111.09543}.

\bibitem[Huang et~al.(2023)Huang, Gupta, Xia, Li, and Chen]{huang2023catastrophicjailbreakopensourcellms}
Yangsibo Huang, Samyak Gupta, Mengzhou Xia, Kai Li, and Danqi Chen.
\newblock Catastrophic jailbreak of open-source llms via exploiting generation, 2023.
\newblock URL \url{https://arxiv.org/abs/2310.06987}.

\bibitem[Huang et~al.(2024)Huang, Wang, Jia, Guo, Juefei-Xu, Zhang, Pu, and Liu]{huang2024semanticguidedpromptorganizationuniversal}
Yihao Huang, Chong Wang, Xiaojun Jia, Qing Guo, Felix Juefei-Xu, Jian Zhang, Geguang Pu, and Yang Liu.
\newblock Semantic-guided prompt organization for universal goal hijacking against llms, 2024.
\newblock URL \url{https://arxiv.org/abs/2405.14189}.

\bibitem[Hughes et~al.(2024)Hughes, Price, Lynch, Schaeffer, Barez, Koyejo, Sleight, Jones, Perez, and Sharma]{hughes2024bestofnjailbreaking}
John Hughes, Sara Price, Aengus Lynch, Rylan Schaeffer, Fazl Barez, Sanmi Koyejo, Henry Sleight, Erik Jones, Ethan Perez, and Mrinank Sharma.
\newblock Best-of-n jailbreaking, 2024.
\newblock URL \url{https://arxiv.org/abs/2412.03556}.

\bibitem[Inan et~al.(2023)Inan, Upasani, Chi, Rungta, Iyer, Mao, Tontchev, Hu, Fuller, Testuggine, and Khabsa]{inan2023llamaguardllmbasedinputoutput}
Hakan Inan, Kartikeya Upasani, Jianfeng Chi, Rashi Rungta, Krithika Iyer, Yuning Mao, Michael Tontchev, Qing Hu, Brian Fuller, Davide Testuggine, and Madian Khabsa.
\newblock Llama guard: Llm-based input-output safeguard for human-ai conversations, 2023.
\newblock URL \url{https://arxiv.org/abs/2312.06674}.

\bibitem[Kim et~al.(2023)Kim, Derakhshan, and Harris]{kim2023robustsafetyclassifierlarge}
Jinhwa Kim, Ali Derakhshan, and Ian~G. Harris.
\newblock Robust safety classifier for large language models: Adversarial prompt shield, 2023.
\newblock URL \url{https://arxiv.org/abs/2311.00172}.

\bibitem[lakera.ai(2024)]{LakeraAI2024}
lakera.ai.
\newblock Lakera-guard, 2024.
\newblock lakera.ai. 2024. Lakera-guard.

\bibitem[Li and Liu(2024)]{li2024injecguardbenchmarkingmitigatingoverdefense}
Hao Li and Xiaogeng Liu.
\newblock Injecguard: Benchmarking and mitigating over-defense in prompt injection guardrail models, 2024.
\newblock URL \url{https://arxiv.org/abs/2410.22770}.

\bibitem[Li et~al.(2024{\natexlab{a}})Li, Dong, Wang, Hu, Zuo, Lin, Qiao, and Shao]{li2024salad}
Lijun Li, Bowen Dong, Ruohui Wang, Xuhao Hu, Wangmeng Zuo, Dahua Lin, Yu~Qiao, and Jing Shao.
\newblock Salad-bench: A hierarchical and comprehensive safety benchmark for large language models.
\newblock \emph{arXiv preprint arXiv:2402.05044}, 2024{\natexlab{a}}.

\bibitem[Li et~al.(2024{\natexlab{b}})Li, Dong, Wang, Hu, Zuo, Lin, Qiao, and Shao]{li2024saladbenchhierarchicalcomprehensivesafety}
Lijun Li, Bowen Dong, Ruohui Wang, Xuhao Hu, Wangmeng Zuo, Dahua Lin, Yu~Qiao, and Jing Shao.
\newblock Salad-bench: A hierarchical and comprehensive safety benchmark for large language models, 2024{\natexlab{b}}.
\newblock URL \url{https://arxiv.org/abs/2402.05044}.

\bibitem[Lin et~al.(2023)Lin, Wang, Tong, Wang, Guo, Wang, and Shang]{lin2023toxicchat}
Zi~Lin, Zihan Wang, Yongqi Tong, Yangkun Wang, Yuxin Guo, Yujia Wang, and Jingbo Shang.
\newblock Toxicchat: Unveiling hidden challenges of toxicity detection in real-world user-ai conversation, 2023.

\bibitem[Liu et~al.(2023)Liu, Lin, Hewitt, Paranjape, Bevilacqua, Petroni, and Liang]{liu2023lostmiddlelanguagemodels}
Nelson~F. Liu, Kevin Lin, John Hewitt, Ashwin Paranjape, Michele Bevilacqua, Fabio Petroni, and Percy Liang.
\newblock Lost in the middle: How language models use long contexts, 2023.
\newblock URL \url{https://arxiv.org/abs/2307.03172}.

\bibitem[Liu et~al.(2024{\natexlab{a}})Liu, Yu, Zhang, Zhang, and Xiao]{liu2024automaticuniversalpromptinjection}
Xiaogeng Liu, Zhiyuan Yu, Yizhe Zhang, Ning Zhang, and Chaowei Xiao.
\newblock Automatic and universal prompt injection attacks against large language models, 2024{\natexlab{a}}.
\newblock URL \url{https://arxiv.org/abs/2403.04957}.

\bibitem[Liu et~al.(2024{\natexlab{b}})Liu, Jia, Geng, Jia, and Gong]{liu2024formalizingbenchmarkingpromptinjection}
Yupei Liu, Yuqi Jia, Runpeng Geng, Jinyuan Jia, and Neil~Zhenqiang Gong.
\newblock Formalizing and benchmarking prompt injection attacks and defenses, 2024{\natexlab{b}}.
\newblock URL \url{https://arxiv.org/abs/2310.12815}.

\bibitem[Meta(2024)]{meta2024}
Meta.
\newblock Promptguard prompt injection guardrail, 2024.
\newblock URL \url{https://www.llama.com/docs/model-cards-and-prompt-formats/prompt-guard}.

\bibitem[Padhi et~al.(2024)Padhi, Nagireddy, Cornacchia, Chaudhury, Pedapati, Dognin, Murugesan, Miehling, Cooper, Fraser, Zizzo, Hameed, Purcell, Desmond, Pan, Ashktorab, Vejsbjerg, Daly, Hind, Geyer, Rawat, Varshney, and Sattigeri]{padhi2024graniteguardian}
Inkit Padhi, Manish Nagireddy, Giandomenico Cornacchia, Subhajit Chaudhury, Tejaswini Pedapati, Pierre Dognin, Keerthiram Murugesan, Erik Miehling, Martín~Santillán Cooper, Kieran Fraser, Giulio Zizzo, Muhammad~Zaid Hameed, Mark Purcell, Michael Desmond, Qian Pan, Zahra Ashktorab, Inge Vejsbjerg, Elizabeth~M. Daly, Michael Hind, Werner Geyer, Ambrish Rawat, Kush~R. Varshney, and Prasanna Sattigeri.
\newblock Granite guardian, 2024.
\newblock URL \url{https://arxiv.org/abs/2412.07724}.

\bibitem[Perez and Ribeiro(2022)]{perez2022ignorepreviouspromptattack}
Fábio Perez and Ian Ribeiro.
\newblock Ignore previous prompt: Attack techniques for language models, 2022.
\newblock URL \url{https://arxiv.org/abs/2211.09527}.

\bibitem[ProtectAI.com(2024{\natexlab{a}})]{deberta-v3-base-prompt-injection-v2}
ProtectAI.com.
\newblock Fine-tuned deberta-v3-base for prompt injection detection, 2024{\natexlab{a}}.
\newblock URL \url{https://huggingface.co/ProtectAI/deberta-v3-base-prompt-injection-v2}.

\bibitem[ProtectAI.com(2024{\natexlab{b}})]{protectai2024}
ProtectAI.com.
\newblock Fine-tuned deberta-v3-base for prompt injection detection, 2024{\natexlab{b}}.
\newblock URL \url{https://huggingface.co/protectai/deberta-v3-base-prompt-injection-v2}.

\bibitem[Qi et~al.(2023)Qi, Zeng, Xie, Chen, Jia, Mittal, and Henderson]{qi2023finetuningalignedlanguagemodels}
Xiangyu Qi, Yi~Zeng, Tinghao Xie, Pin-Yu Chen, Ruoxi Jia, Prateek Mittal, and Peter Henderson.
\newblock Fine-tuning aligned language models compromises safety, even when users do not intend to!, 2023.
\newblock URL \url{https://arxiv.org/abs/2310.03693}.

\bibitem[Rebedea et~al.(2023)Rebedea, Dinu, Sreedhar, Parisien, and Cohen]{rebedea2023nemoguardrailstoolkitcontrollable}
Traian Rebedea, Razvan Dinu, Makesh Sreedhar, Christopher Parisien, and Jonathan Cohen.
\newblock Nemo guardrails: A toolkit for controllable and safe llm applications with programmable rails, 2023.
\newblock URL \url{https://arxiv.org/abs/2310.10501}.

\bibitem[Samvelyan et~al.(2024)Samvelyan, Raparthy, Lupu, Hambro, Markosyan, Bhatt, Mao, Jiang, Parker-Holder, Foerster, Rocktäschel, and Raileanu]{samvelyan2024rainbowteamingopenendedgeneration}
Mikayel Samvelyan, Sharath~Chandra Raparthy, Andrei Lupu, Eric Hambro, Aram~H. Markosyan, Manish Bhatt, Yuning Mao, Minqi Jiang, Jack Parker-Holder, Jakob Foerster, Tim Rocktäschel, and Roberta Raileanu.
\newblock Rainbow teaming: Open-ended generation of diverse adversarial prompts, 2024.
\newblock URL \url{https://arxiv.org/abs/2402.16822}.

\bibitem[Sanh et~al.(2020)Sanh, Debut, Chaumond, and Wolf]{sanh2020distilbertdistilledversionbert}
Victor Sanh, Lysandre Debut, Julien Chaumond, and Thomas Wolf.
\newblock Distilbert, a distilled version of bert: smaller, faster, cheaper and lighter, 2020.
\newblock URL \url{https://arxiv.org/abs/1910.01108}.

\bibitem[Shen et~al.(2024)Shen, Chen, Backes, Shen, and Zhang]{SCBSZ24}
Xinyue Shen, Zeyuan Chen, Michael Backes, Yun Shen, and Yang Zhang.
\newblock {``Do Anything Now'': Characterizing and Evaluating In-The-Wild Jailbreak Prompts on Large Language Models}.
\newblock In \emph{{ACM SIGSAC Conference on Computer and Communications Security (CCS)}}. ACM, 2024.

\bibitem[Shi et~al.(2024)Shi, Yuan, Liu, Huang, Zhou, Sun, and Gong]{shi2024optimizationbasedpromptinjectionattack}
Jiawen Shi, Zenghui Yuan, Yinuo Liu, Yue Huang, Pan Zhou, Lichao Sun, and Neil~Zhenqiang Gong.
\newblock Optimization-based prompt injection attack to llm-as-a-judge, 2024.
\newblock URL \url{https://arxiv.org/abs/2403.17710}.

\bibitem[Team(2024)]{qwen2024}
Qwen Team.
\newblock A party of foundation models, 2024.
\newblock URL \url{https://qwenlm.github.io/blog/qwen2.5}.

\bibitem[Tedeschi et~al.(2024)Tedeschi, Friedrich, Schramowski, Kersting, Navigli, Nguyen, and Li]{tedeschi2024alert}
Simone Tedeschi, Felix Friedrich, Patrick Schramowski, Kristian Kersting, Roberto Navigli, Huu Nguyen, and Bo~Li.
\newblock Alert: A comprehensive benchmark for assessing large language models' safety through red teaming, 2024.

\bibitem[Toyer et~al.(2023)Toyer, Watkins, Mendes, Svegliato, Bailey, Wang, Ong, Elmaaroufi, Abbeel, Darrell, Ritter, and Russell]{toyer2023tensortrustinterpretableprompt}
Sam Toyer, Olivia Watkins, Ethan~Adrian Mendes, Justin Svegliato, Luke Bailey, Tiffany Wang, Isaac Ong, Karim Elmaaroufi, Pieter Abbeel, Trevor Darrell, Alan Ritter, and Stuart Russell.
\newblock Tensor trust: Interpretable prompt injection attacks from an online game, 2023.
\newblock URL \url{https://arxiv.org/abs/2311.01011}.

\bibitem[Wang et~al.(2023)Wang, Freire, Zhang, Wei, Goncalves, Kostakos, Sarsenbayeva, Schneegass, Bozzon, and Niforatos]{wang2023safeguardingcrowdsourcingsurveyschatgpt}
Chaofan Wang, Samuel~Kernan Freire, Mo~Zhang, Jing Wei, Jorge Goncalves, Vassilis Kostakos, Zhanna Sarsenbayeva, Christina Schneegass, Alessandro Bozzon, and Evangelos Niforatos.
\newblock Safeguarding crowdsourcing surveys from chatgpt with prompt injection, 2023.
\newblock URL \url{https://arxiv.org/abs/2306.08833}.

\bibitem[Wang et~al.(2024)Wang, Shi, Bai, and Hsieh]{wang2024defendingllmsjailbreakingattacks}
Yihan Wang, Zhouxing Shi, Andrew Bai, and Cho-Jui Hsieh.
\newblock Defending llms against jailbreaking attacks via backtranslation, 2024.
\newblock URL \url{https://arxiv.org/abs/2402.16459}.

\bibitem[Warner et~al.(2024)Warner, Chaffin, Clavié, Weller, Hallström, Taghadouini, Gallagher, Biswas, Ladhak, Aarsen, Cooper, Adams, Howard, and Poli]{warner2024smarterbetterfasterlonger}
Benjamin Warner, Antoine Chaffin, Benjamin Clavié, Orion Weller, Oskar Hallström, Said Taghadouini, Alexis Gallagher, Raja Biswas, Faisal Ladhak, Tom Aarsen, Nathan Cooper, Griffin Adams, Jeremy Howard, and Iacopo Poli.
\newblock Smarter, better, faster, longer: A modern bidirectional encoder for fast, memory efficient, and long context finetuning and inference, 2024.
\newblock URL \url{https://arxiv.org/abs/2412.13663}.

\bibitem[Wei et~al.(2023)Wei, Haghtalab, and Steinhardt]{wei2023jailbrokendoesllmsafety}
Alexander Wei, Nika Haghtalab, and Jacob Steinhardt.
\newblock Jailbroken: How does llm safety training fail?, 2023.
\newblock URL \url{https://arxiv.org/abs/2307.02483}.

\bibitem[Wolf et~al.(2020)Wolf, Debut, Sanh, Chaumond, Delangue, Moi, Cistac, Rault, Louf, Funtowicz, Davison, Shleifer, von Platen, Ma, Jernite, Plu, Xu, Scao, Gugger, Drame, Lhoest, and Rush]{wolf-etal-2020-transformers}
Thomas Wolf, Lysandre Debut, Victor Sanh, Julien Chaumond, Clement Delangue, Anthony Moi, Pierric Cistac, Tim Rault, Rémi Louf, Morgan Funtowicz, Joe Davison, Sam Shleifer, Patrick von Platen, Clara Ma, Yacine Jernite, Julien Plu, Canwen Xu, Teven~Le Scao, Sylvain Gugger, Mariama Drame, Quentin Lhoest, and Alexander~M. Rush.
\newblock Transformers: State-of-the-art natural language processing.
\newblock In \emph{Proceedings of the 2020 Conference on Empirical Methods in Natural Language Processing: System Demonstrations}, pages 38--45, Online, October 2020. Association for Computational Linguistics.
\newblock URL \url{https://www.aclweb.org/anthology/2020.emnlp-demos.6}.

\bibitem[Wolf et~al.(2024)Wolf, Wies, Avnery, Levine, and Shashua]{wolf2024fundamentallimitationsalignmentlarge}
Yotam Wolf, Noam Wies, Oshri Avnery, Yoav Levine, and Amnon Shashua.
\newblock Fundamental limitations of alignment in large language models, 2024.
\newblock URL \url{https://arxiv.org/abs/2304.11082}.

\bibitem[Yi et~al.(2023)Yi, Xie, Zhu, Hines, Kiciman, Sun, Xie, and Wu]{yi2023benchmarking}
Jingwei Yi, Yueqi Xie, Bin Zhu, Keegan Hines, Emre Kiciman, Guangzhong Sun, Xing Xie, and Fangzhao Wu.
\newblock Benchmarking and defending against indirect prompt injection attacks on large language models.
\newblock \emph{arXiv preprint arXiv:2312.14197}, 2023.

\bibitem[Yu et~al.(2024)Yu, Lin, Yu, and Xing]{299691}
Jiahao Yu, Xingwei Lin, Zheng Yu, and Xinyu Xing.
\newblock {LLM-Fuzzer}: Scaling assessment of large language model jailbreaks.
\newblock In \emph{33rd USENIX Security Symposium (USENIX Security 24)}, pages 4657--4674, Philadelphia, PA, August 2024. USENIX Association.
\newblock ISBN 978-1-939133-44-1.
\newblock URL \url{https://www.usenix.org/conference/usenixsecurity24/presentation/yu-jiahao}.

\bibitem[Zeng et~al.(2024)Zeng, Liu, Mullins, Peran, Fernandez, Harkous, Narasimhan, Proud, Kumar, Radharapu, Sturman, and Wahltinez]{zeng2024shieldgemmagenerativeaicontent}
Wenjun Zeng, Yuchi Liu, Ryan Mullins, Ludovic Peran, Joe Fernandez, Hamza Harkous, Karthik Narasimhan, Drew Proud, Piyush Kumar, Bhaktipriya Radharapu, Olivia Sturman, and Oscar Wahltinez.
\newblock Shieldgemma: Generative ai content moderation based on gemma, 2024.
\newblock URL \url{https://arxiv.org/abs/2407.21772}.

\bibitem[Zhan et~al.(2024)Zhan, Liang, Ying, and Kang]{zhan2024injecagentbenchmarkingindirectprompt}
Qiusi Zhan, Zhixiang Liang, Zifan Ying, and Daniel Kang.
\newblock Injecagent: Benchmarking indirect prompt injections in tool-integrated large language model agents, 2024.
\newblock URL \url{https://arxiv.org/abs/2403.02691}.

\bibitem[Zhou et~al.(2025)Zhou, Han, Zhuang, Guo, Liang, Bao, and Zhang]{zhou2025defendingjailbreakpromptsincontext}
Yujun Zhou, Yufei Han, Haomin Zhuang, Kehan Guo, Zhenwen Liang, Hongyan Bao, and Xiangliang Zhang.
\newblock Defending jailbreak prompts via in-context adversarial game, 2025.
\newblock URL \url{https://arxiv.org/abs/2402.13148}.

\bibitem[Zou et~al.(2023)Zou, Wang, Carlini, Nasr, Kolter, and Fredrikson]{zou2023universaltransferableadversarialattacks}
Andy Zou, Zifan Wang, Nicholas Carlini, Milad Nasr, J.~Zico Kolter, and Matt Fredrikson.
\newblock Universal and transferable adversarial attacks on aligned language models, 2023.
\newblock URL \url{https://arxiv.org/abs/2307.15043}.

\bibitem[Zou et~al.(2024)Zou, Phan, Wang, Duenas, Lin, Andriushchenko, Wang, Kolter, Fredrikson, and Hendrycks]{zou2024improvingalignmentrobustnesscircuit}
Andy Zou, Long Phan, Justin Wang, Derek Duenas, Maxwell Lin, Maksym Andriushchenko, Rowan Wang, Zico Kolter, Matt Fredrikson, and Dan Hendrycks.
\newblock Improving alignment and robustness with circuit breakers, 2024.
\newblock URL \url{https://arxiv.org/abs/2406.04313}.

\end{thebibliography}

\end{document}